\documentclass[lettersize,journal]{IEEEtran}

\usepackage{amsmath,amsfonts}
\usepackage{algorithmic}
\usepackage{array}
\usepackage[caption=false,font=normalsize,labelfont=sf,textfont=sf]{subfig}
\usepackage{textcomp}
\usepackage{stfloats}
\usepackage{url}
\usepackage{verbatim}
\usepackage{graphicx}
\usepackage{cleveref}
\def\BibTeX{{\rm B\kern-.05em{\sc i\kern-.025em b}\kern-.08em
    T\kern-.1667em\lower.7ex\hbox{E}\kern-.125emX}}
\usepackage{balance}
\usepackage{bm}
\usepackage{xcolor}

\usepackage{algorithm}
\usepackage{algorithmic}
\usepackage{enumitem}
\usepackage[normalem]{ulem}
\usepackage{soul}
\usepackage{balance}

\begin{document}
\title{An Efficient Representation of Whole-body Model Predictive Control for Online Compliant Dual-arm Mobile Manipulation}
\author{Wenqian Du$^{{1},{2}}$, Ran Long$^{{1}}$, Jo\~{a}o Moura$^{{1},{2}}$, Jiayi Wang$^{{1}}$, Saeid Samadi$^{1}$, Sethu Vijayakumar$^{{1},{2}}$
\thanks{Manuscript created in June, 2024;  }
		\thanks{$^{1}$ Authors are with the School of Informatics, The University of Edinburgh, Edinburgh, U.K. 
	wdu2@ed.ac.uk}%
\thanks{$^{2}$ Authors are with The Alan Turing Institute, London, U.K.}
\thanks{Digital Object Identifier (DOI): see top of this page.}
}

\markboth{IEEE Transactions on Robotics, VOL. XX, NO. XX, June~2024}
{How to Use the IEEEtran \LaTeX \ Templates}

\maketitle

\begin{abstract}
Dual-arm mobile manipulators can transport and manipulate large-size objects with simple end-effectors. To interact with dynamic environments with strict safety and compliance requirements, 
achieving whole-body motion planning online while meeting various hard constraints for such highly redundant mobile manipulators poses a significant challenge.
We tackle this challenge by presenting an efficient representation of whole-body motion trajectories within our bilevel model-based predictive control (MPC) framework.
We utilize B\'ezier-curve parameterization to represent the optimized collision-free trajectories of two collaborating end-effectors in the first MPC, facilitating fast long-horizon object-oriented motion planning in $SE(3)$ while considering approximated feasibility constraints. This approach is further applied to parameterize whole-body trajectories in the second MPC for whole-body motion generation with predictive admittance control in a relatively short horizon while satisfying whole-body hard constraints. This representation enables two MPCs with continuous properties, thereby avoiding inaccurate model-state transition and dense decision-variable settings in existing MPCs using the discretization method. It strengthens the online execution of the bilevel MPC framework in high-dimensional space and facilitates the generation of consistent commands for our hybrid position/velocity-controlled robot. 
The simulation comparisons and real-world experiments demonstrate the efficiency and robustness of this approach in various scenarios for static and dynamic obstacle avoidance, and compliant interaction control with the manipulated object and external disturbances. 
\end{abstract}

\begin{IEEEkeywords}
Bilevel model-based predictive control, B\'ezier Curve, dual-arm mobile manipulation, online task-space motion planning, online whole-body trajectory optimization, predictive admittance Control
\end{IEEEkeywords}

\section{Introduction}\label{intro}
\IEEEPARstart{M}{obile} manipulators have a wide range of real-world applications, such as object transportation in unmanned warehouses or manufacturing industries. 
In particular, dual-arm manipulators have better manipulation capability and can handle large objects using simpler end-effectors. 

\begin{figure}[t]
\centering
\includegraphics[height=9.25cm]{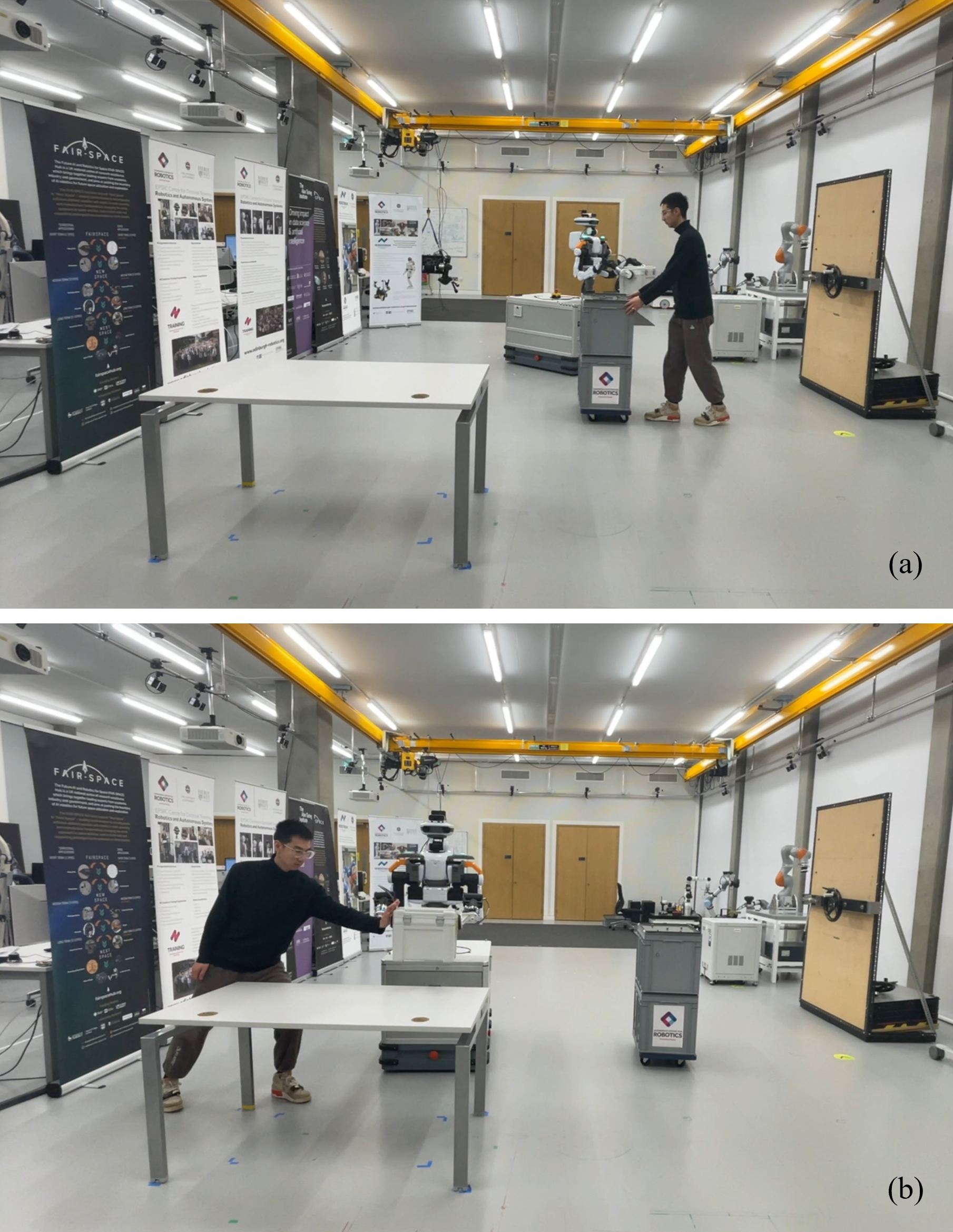}
\caption{Dual-arm mobile manipulation while avoiding dynamic obstacles and conducting push-recovery. The experimental video can be accessed at https://youtu.be/CEoeLRtpyHw.}
\label{EVA}
\end{figure}

To enable automation in the real world, the robot should have the capability of avoiding dynamic obstacles in changing environments where the robot base and arms are subject to potential collisions as shown in Fig. \ref{EVA}(a). To handle dynamic obstacle avoidance, long-horizon planning is advantageous for mobile robots, enabling the generation of smooth collision-free trajectories. To transport an object with two end-effectors with solid and safe contacts, the robot should interact compliantly with the object in the moving process. To ensure safety when collaborating with humans, the robot should exhibit online compliance with external force disturbances as shown in Fig. \ref{EVA}(b). By only providing the object goal, these criteria make it necessary for the robot to re-plan collaborative motion between two end-effectors and whole-body compliant motion online considering various constraints, which is often formulated as trajectory optimization (TO) problem \cite{hrovat2012development}. 
However, for the dual-arm mobile manipulator with a high degree of freedom (DOF), the complex robot model and interaction control model induce high non-linearity into the TO problem \cite{spahn2021coupled}. 
Consequently, this high-dimensional and non-convex problem is difficult to solve online when long-horizon whole-body motion planning is required and various hard constraints should be satisfied. Without a motion reference, this holistic non-convex problem can be easily stuck at bad local minima \cite{jelavic2023lstp}. 

In this paper, we present a novel whole-body motion planning approach for highly redundant mobile manipulators. 
The key strength of our method lies in its fast computation speed, enabling robust online planning of both task-space and whole-body joint trajectories.

The remainder of this paper is organized as follows. In Section \ref{related}, we introduce the related works while presenting our concepts and contributions of online whole-body motion planning for dual-arm mobile manipulation. Section \ref{bilevel} constitutes the main contribution of this paper and presents the detailed methods. In Section \ref{demo}, we perform several simulations and experiments to show the efficiency and robustness of our approach. Section \ref{conclusion} closes this paper with the conclusion.

\section{Related Work}\label{related}

To mitigate the slow computation of holistic whole-body motion planning, a divide-and-conquer strategy is often applied in the literature to divide the holistic problem into two-stage sub-problems. This strategy has been applied for planning the whole-body motion of manipulators \cite{leu2022long, stouraitis2020online}, humanoid \cite{tonneau2018efficient} and quadruped \cite{grandia2023perceptive} robots, and legged-wheeled robots \cite{jelavic2023lstp}. 
Grandia et al. \cite{grandia2023perceptive} shows that the pre-calculated quadruped-torso motion reference in the first stage can speed up the convergence of the whole-body motion planning. 
The adaptive motion reference generated online from the first stage can function as attractors mitigating the second-stage tendency to converge towards bad local minima \cite{jelavic2023lstp}. 

In this paper, we also employ the two-stage strategy to reduce the computational complexity of each stage. Based on TO to integrate various equality and inequality constraints, we build a bilevel MPC framework for whole-body motion planning of our dual-arm mobile manipulator. 
The first MPC is implemented for task-space motion planning, and its output serves as the motion reference of the second MPC which generates whole-body motion to track the motion reference. However, with only the two-stage strategy, it is still challenging to execute each MPC online. In addition to the high-dimensional and non-convex problem described above, the first MPC needs to address coupling motion planning of two end-effectors in both translational and rotational motion, and the second MPC needs to handle the whole-body complex model and incorporate the interaction control model to perform local force adaption. 
\subsection{TO-based Reactive Task-space Motion Planning}
Fast task-space motion replanning based on TO has been extensively proposed for mobile robots to generate collision-free paths online \cite{zhou2020ego, qian2016motion, kolter2009task, bucki2020rectangular}.
Compared to TO that builds cost functions and constraints on discretized points of trajectories \cite{rosmann2017integrated, szmuk2018real}, the integration of splines with polyhedral outer representations in task-space TO enables the satisfaction of safety constraints at all points on the trajectory \cite{zhou2019robust, tordesillas2021mader}. In particular, Tordesillas et al. \cite{tordesillas2021mader} employed this method to achieve the significant efficiency of multi-agents in conducting collision avoidance among multiple static and dynamic obstacles. 
However, the aforementioned works primarily focus on generating task-space point-position trajectories to handle obstacle avoidance. 

In our work, to handle dual-arm manipulation with collaborative motion between end-effectors, we need to optimize the end-effectors' translational and rotational motions simultaneously. However, TO-based online coupled motion re-planning (from the robot's current state to a goal) in $SE(E)$ still remains an open issue. Only a few studies fall in this remit.
Zhou et al. \cite{zhou2021raptor} develop a two-step yaw angle planning for a quadrotor to facilitate active exploration of unknown spaces. However, the position and orientation planning are solved separately using two TOs. 
Nikhil et al. \cite{potdar2020online} plan the position trajectory for a payload micro aerial vehicle while minimizing its roll and pitch angles and maintaining a constant yaw. 
Stouraitis et al. \cite{stouraitis2020online} utilize a discretization approach to optimize the pose trajectory of an object to be manipulated, considering both its position and Euler angle-based orientation.
However, Euler angle representation is susceptive to singularity, and since interpolation on Euler angles ignores interdependencies among rotation axes, it may lead to unexpected behaviors that are neither optimal nor intuitively correct, as detailed in \cite{dam1998quaternions, allmendinger2018coordinate}. In contrast, quaternion representation avoids these problems \cite{dam1998quaternions}.
However, the integration of quaternions into TO for optimizing the intermediate points between initial and final orientation states is not straightforward in terms of computation costs while satisfying unit-quaternion constraints, especially leveraging the efficient spline-based method. 
In this paper, we explore an efficient way to incorporate quaternion representation in spline-based TO to achieve fast coupled motion planning of two cooperating end-effectors between $\mathbb{R}^3$ and $SO(3)$. 

\subsection{Whole-body Motion Generation for Mobile Manipulation}
Compared with independently operating the mobile base and the manipulator in a sequential order \cite{rehman2016towards}, coupling whole-body motion reduces total operational time considerably and leads to smoother and more consistent motions between the mobile base and manipulator \cite{thakar2018towards, burgess2024reactive}. 

The whole-body motion generation of mobile manipulation, employing a receding-horizon TO based on robot models, 
outperforms inverse kinematics-based whole-body instantaneous control, as the latter struggles to avoid collisions with the environment with a much lower success rate \cite{mittal2022articulated}. 
Ide et al. \cite{ide2011real} develop a real-time MPC controller for a 5-DOF mobile manipulator without collision avoidance. 
Avanzini et al. \cite{avanzini2015constraint} generate the whole-body motion of an 8-DOF mobile manipulator for dynamic obstacle avoidance. This is extended to non-holonomic bases with object detection \cite{avanzini2018constrained}. 
Spahn et al. \cite{spahn2021coupled} reduce the number of obstacle-avoidance constraints in MPC for a 10-DOF mobile manipulator by modeling the free space as convex polyhedrons. 
In these works, the MPCs are designed to track references of the base and arm-joint motion to constitute convex cost functions in order to speed up the MPC computation; however, this needs pre-transformation from end-effector motion to the joint motion at designed time knots. 

The more straightforward way is to track the end-effector motion reference directly by integrating the kinematics model in the tracking cost function to optimize the joint motion, however, this seriously increases the MPC computation time \cite{avanzini2015constraint}. 
The works in \cite{pankert2020perceptive, minniti2019whole} optimize the whole-body motion by tracking the end-effector motion references and they utilize relaxed constraints to achieve online performance, however, this compromises the safety requirements of the robots.
In addition, since the MPC computation time scales with robot DOF \cite{spahn2021coupled}, executing high-dimensional and non-convex MPC online for high-DOF robots presents significant challenges, especially for our dual-arm mobile manipulator. 

In whole-body motion generation, compliant interaction control for safety is also required when a robot interacts with humans and the environment. 
Impedance/admittance controllers \cite{hogan1984impedance} are recently incorporated into MPC \cite{matschek2017force, gold2022model, kleff2022introducing, bednarczyk2020model, kazim2018combined} for industrial manipulators. The integrated interaction force control in MPC ensures that the combined trajectory of the motion response curve and the planned trajectory satisfies constraints such as joint/velocity limits, safety requirements and obstacle avoidance \cite{bednarczyk2020model}. MPC-based admittance control \cite{ wahrburg2016mpc} shows superior contact stability compared to standard instantaneous admittance control. Since these works only concentrate on manipulators with few DOFs, 
the integration of predictive admittance control for whole-body motion generation poses considerable difficulties for high-DOF robots due to the increased complexity in mapping motion responses at end-effectors to joint motions while considering whole-body hard constraints. 
This creates significant computational challenges for executing the whole-body MPC with predictive force control, especially for our dual-arm mobile manipulator with high DOFs. 

\subsection{Model-based Predictive Control for High-DOF Robots}\label{MPCs}
MPC performance is sensitive to computational time which limit its application to high-dimensional configuration spaces, especially when a relatively long horizon is expected. 
Dai et al. \cite{dai2014whole} combine centroidal dynamics and whole-body kinematics models to generate whole-body motion for a quadruped and a humanoid, respectively, but it takes seconds to converge through common nonlinear optimization solvers. To speed up TO, Bernardo et al. \cite{aceituno2017simultaneous} formulate the whole-body motion generation of a quadruped as a convex TO using a simplified dynamics model and approximated whole-body constraints. However, its computation is still not fast enough for online execution, and the authors find that the simplified model does not generalize well for more dynamic gaits since the model cannot induce limits on angular momentum. The simplified model is also applied in \cite{henze2014posture} to achieve online motion generation of a humanoid, without incorporating joint-limit constraints. 
In addition, Yu et al. \cite{yu2023modeling} apply linearization on the TO of a wheeled bipedal robot, and employ quadratic programming \cite{ferreau2014qpoases} to solve the linear MPC problem online, however, linearization error is induced and it is sensitive to system changes. 

Another category of works develops specialized solvers that are applied to exploit the sparsity and variants of differential dynamic programming (DDP) using Gauss-Newton approximation, such as Sequential Linear Quadratic \cite{farshidian2017efficient}, iterative Linear Quadratic Gaussian \cite{tassa2012synthesis}, Feasibility-driven DDP \cite{mastalli2020crocoddyl}. The works in \cite{chiu2022collision, bjelonic2022offline, sleiman2021unified} use the relaxed barrier function \cite{feller2016relaxed} in MPC to absorb inequality constraints into the cost function, to achieve online MPC execution. However, methods using DDP and barrier functions apply soft constraints, which may struggle to ensure safe interaction with the environment. 

In the above MPC works for whole-body motion generation, the discretization approach is used to formulate the MPCs in which model states of the robot configuration, velocity, and acceleration, force/torque are treated as different decision variables. These variables are interconnected through whole-body model-state transition, such as the Euler approximation and Runge-Kutta methods. However, this model-state transition is not accurate, particularly when the time interval between two consecutive time knots is large. This inaccuracy results in inconsistencies among different states, such as inconsistent joint position and velocity commands for our hybrid position/velocity-controlled robot. This inaccurate model-state transition also impacts the feasible solution manifold, making the discretized MPC extremely sensitive to horizon lengths.
To improve the transition accuracy and enhance the motion-tracking performance, dense discretization is deployed in the above MPC works. This leads to an excessive number of decision variables at dense time knots, therefore limiting the MPC computation speed. 
For our dual-arm mobile manipulator, the high-dimensional joint and task space further enlarge the number of decision variables. 

\subsection{Contributions}
In this paper, our objective is to achieve whole-body motion planning online. In our bilevel MPC framework, we construct the task-space and whole-body MPCs using hard constraints to ensure stringent safety requirements, while also preserving the model accuracy and handling the nonlinear TO without linearization. To speed up their computation, we integrate an efficient representation aimed at decreasing the complexity of each TO problem.
Our approach utilizes a specific spline known as B\'ezier curves, which embody elegant properties, including curve fitting \cite{qian2016motion}, convex hull \cite{tordesillas2021mader}, continuity \cite{fernbach2020c, kicki2023fast}, and smooth interpolation \cite{dam1998quaternions}. 
To the best of our knowledge, our research is the first to assess the deployment of B\'ezier-curve representation's effect on high-DOF robots within the realm of online MPC-based motion planning. 
We summarize the key contributions of this paper as follows:
\setlist[itemize]{leftmargin=*}
\begin{itemize}
\item We construct a bilevel-MPC framework for dual-arm mobile manipulation. The first MPC computes a long-horizon trajectory for two collaborating end-effectors while considering an approximated feasibility constraint of the robot. The second MPC optimizes the whole-body motion over a relatively short horizon that simultaneously follows and corrects the end-effectors' trajectories accounting for the whole-body motion constraints and local force adaptation. Both MPCs run in a closed loop to enable robustness to the changing environment.
\item We incorporate the B\'ezier-curve representation of the high-dimensional trajectories in the two MPCs to avoid inaccurate model-state transition and excessive decision variables. This accelerates their computation speeds and enables stable computation times across different horizon lengths. Specifically, with this representation, we achieve three goals: 
\begin{itemize}[label=$\star$]
\item This work presents a novel use of B\'ezier-curve parameterization within the first MPC for achieving fast and adaptive object-oriented \textit{task-space motion planning} in $SE(3)$, particularly through the coupled motion planning of position trajectories in $\mathbb{R}^3$ and efficient quaternion trajectories in $SO(3)$ of two collaborating end-effectors.
\item This is the first work of its type leveraging this efficient representation to transcribe \textit{joint-space trajectories} that satisfy whole-body hard constraints to realize online whole-body motion generation in a high-DOF robot. This facilitates the generation of consistent motion commands for our hybrid position/velocity-controlled robot. 
\item To enhance efficient reaction to external perturbances, 
with this representation, this work also achieves \textit{online predictive admittance control} for a high-DOF robot which is incorporated into the whole-body MPC. 
\end{itemize}
\item Finally, we carry out comparative simulations between discretized MPC formulations and our method, and perform real-world experiments, to demonstrate the efficiency and robustness of our approach. 
\end{itemize}

\begin{figure*}[t]
\centering
\includegraphics[height=6.3cm]{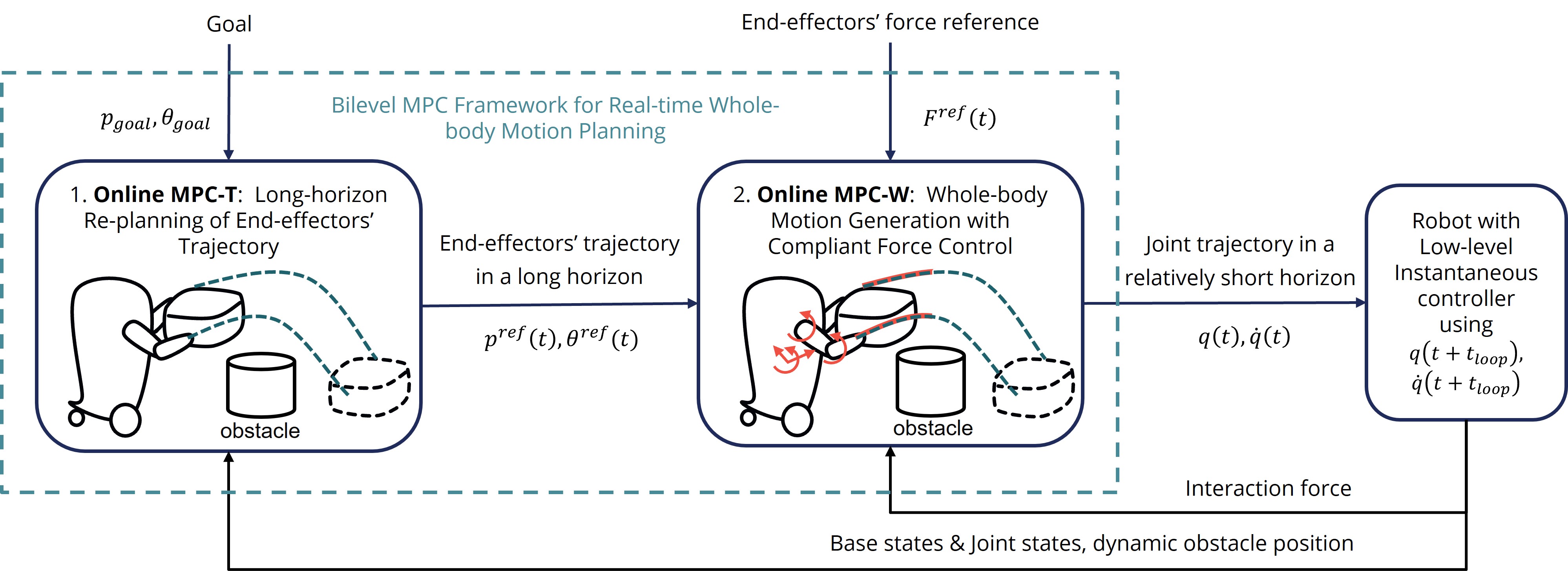}
\caption{Bilevel MPC framework of whole-body compliant motion generation using model-based predictive control. MPC-T in the first stage is introduced in Section \ref{subsec:stage1} and MPC-W in the second stage is presented in Section \ref{subsec:stage2}.}
\label{framework}
\end{figure*}

\section{Bilevel Motion Planning Framework for Dual-arm Mobile Manipulation} \label{bilevel}

\subsection{Bilevel MPC Framework}
The bilevel MPC framework is shown in Fig. \ref{framework}. 
The first MPC relates to object-centric motion planning. Provided with an object's position and orientation goal ($\bm{p}_{goal}, \bm{\theta}_{goal}$), the robot uses its two end-effectors to handle this task-oriented mission. To generate the end-effector's translational $\bm{p}(t)$ and quaternion-based rotational $\bm{\theta}(t)$ trajectories over a long horizon, the first MPC is built in the task space, referred to as MPC-T. The loss function includes the minimization of curve length, velocity and acceleration at defined time knots, and other specific costs for dual-arm manipulation. The constraints include the beginning and goal states as well as collision avoidance.
Approximated base motion limits should be considered in the first stage to ensure its output is either feasible or closely approaching feasibility for the subsequent whole-body motion planning stage. Then the outputs serve as the motion reference of the second stage. The details of the first stage are introduced in \Cref{subsec:stage1}. 

The second MPC termed MPC-W relates to whole-body motion generation over a relatively short horizon. Since the applied dual-arm mobile manipulator is a position/velocity-controlled robot and compliant interaction is required, 
our objective is that
MPC-W optimizes trajectories of joint positions and velocities ($\bm{q}(t), \dot{\bm{q}}(t)$) as well as motion-response trajectories based on predictive admittance control at two end-effectors. The motion responses are used for local force adaption by updating the motion reference from the first stage.
In addition to incorporating loss functions with similar functionalities as those in the first stage, MPC-W tracks the motion reference obtained from the first stage ($\bm{p}^{ref}(t), \bm{\theta}^{ref}(t)$) and provided force references $\bm{F}^{ref}(t)$ at end-effectors. The constraints incorporate the start states, kinematics model, admittance control model, collision avoidance, and joint limits. After MPC-W is solved, the upper body joints with position control receive the generated joint-position commands, while the mobile base under velocity control follows the generated joint-velocity commands. The details of the second stage are introduced in \Cref{subsec:stage2}. 

The bilevel interplay between two MPCs happens within each control loop. MPC-T provides a task-space motion reference to speed up the convergence of MPC-W. 
Since MPC-T does not consider whole-body constraints,
the feasibility of the planned trajectory is not guaranteed. 
MPC-W refines the time knots over a relatively short horizon. With force tracking facilitated by the integrated admittance control and consideration of whole-body constraints, MPC-W generates a whole-body feasible trajectory to follow and correct the end-effectors' motion reference simultaneously. The output of the MPC-W is sent to the robot and the robot moves to a new robot state, which serves as the new inputs of MPC-T and MPC-W in the next control loop. 

\begin{figure*}[t]
\centering
\includegraphics[height=5cm]{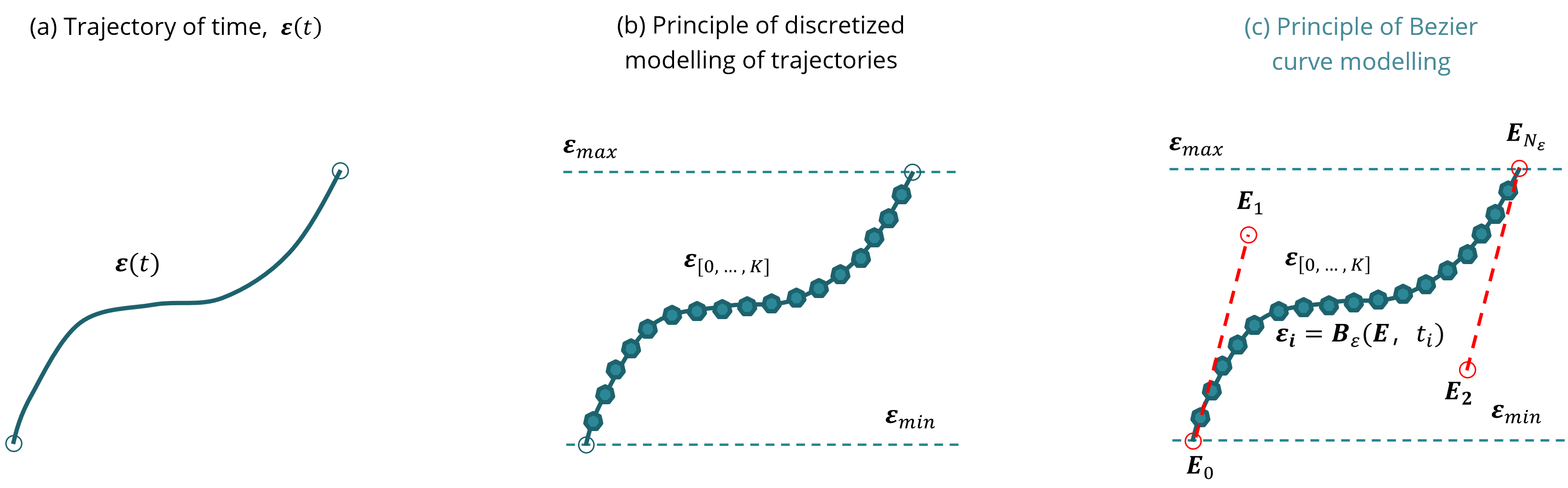}
\caption{(a) Robot state trajectory of time $\bm{\epsilon}(t)$. (b) Discretized robot states at sampling time knots $\bm{\epsilon}_{[0 \cdots K]}$ (c) Concept of B\'ezier-curve transcription. The robot state $\bm{\epsilon}_i$ at $i$'s sampling time knot $\bar{t}_i$ is represented by few control points $\bm{E}$ on a B\'ezier curve $\bm{B}(\bm{E}, \bar{t}_i)$. 
}
\label{concept}
\end{figure*}

\subsection{B\'ezier-curve Representation of Bilevel MPC Framework}\label{bezier_concept}
In the bilevel-MPC framework, each stage TO can be generally built with the continuous formulation as follows,
\begin{equation}
\begin{aligned}
\min_{\bm{\epsilon}(t)} \  &\int_{0}^{T} L_{\epsilon}(\bm{\epsilon}(t)), \\
s.t. \ &g_{\epsilon}(\bm{\epsilon}(t))\leq 0, \ \forall t\in [0, T],   
\end{aligned}
\end{equation}
where $\bm{\epsilon}(t)$ represents the optimized trajectory shown in Fig. \ref{concept}(a). $L_{\epsilon}(\cdot)$ represents the loss function, $g_{\epsilon}(\cdot)$ represents equality and inequality constraints. $T$ is the time horizon. 
To solve this TO, the continuous problem is often formulated as a conventional MPC that provides a numerical solution by discretizing the continuous curve at $K+1$ time knots (shown in Fig. \ref{concept}(b)), 
\begin{equation}\label{discretized_MPC}
\begin{aligned}
\min_{\bm{\epsilon}_{[0, \hdots, K]}} \  &\sum_{i=0}^{K} L_{\epsilon}(\bm{\epsilon}_i), \\
s.t. \ &g_{\epsilon}(\bm{\epsilon}_i)\leq 0, \ \forall i\in [0, K],   
\end{aligned}
\end{equation}
where $\bm{\epsilon}_i$ is the optimized state at $i$-th time knot. Taking MPC-W as an example, 
joint position $\bm{q}$ and velocity $\dot{\bm{q}}$ are both optimized. With this discretization approach, the equality constraint of model-state transition is required to link $\bm{q}$ and $\dot{\bm{q}}$. However, due to the inaccurate transition, the optimized $\bm{q}$ and $\dot{\bm{q}}$ are not consistent with respect to time, which influences the tracking performance of MPC-W for our hybrid position/velocity-controlled robot. 
Since MPC-W also constrains the joint position and velocity limits, the inaccurate transition also influences the feasible solution manifold, which negatively influences the MPC computation time. 
Dense $K$ time knots can increase transition accuracy, but slow down the MPC convergence speed. The performance of this classical approach is shown in Section \ref{demo}. 

To speed up the two MPCs, we aim to reduce the complexity of each MPC effectively. Instead of applying discretization on the state curve $\bm{\epsilon}(t)$, we transcribe all trajectories needed by the two MPCs as B\'ezier curves, including the trajectories of the end-effectors' motion in $SE(3)$, joint motion as well as the motion responses from the integrated admittance control. Rather than approximating the model state transition inaccurately, this representation directly links position, velocity, and acceleration states with respect to time. 

A B\'ezier curve is parameterized by its control points which constitute the convex hull of the whole curve \cite{fernbach2020c}. 
At each time knot, the robot state is the function of the control points.
In Euclidean space, a B\'ezier curve has the following formulation, 
\begin{equation}\label{bezier_general}
	\bm{\epsilon}(t) = \bm{B}_\epsilon(\bm{E}, \ \bar{t}) = \sum_{j=0}^{N_\epsilon} \frac{N_\epsilon!}{j!(N_\epsilon-j)!}\bar{t}^j(1-\bar{t})^{N_\epsilon-j} \bm{E}_{j}, 
\end{equation}
where 
$\bm{B}_\epsilon(\cdot)$ indicates a B\'ezier curve, and its subscript $\epsilon$ indicates the variable parameterized by the B\'ezier-curve formulation. $\bm{E} = [\bm{E}_0 \ \cdots \ \bm{E}_{N_\epsilon}]\in \mathbb{R}^{n_{\epsilon}\times (N_\epsilon+1)}$ denotes the control points of the B\'ezier curve, $N_\epsilon + 1$ is control point number, $j$ is the control point index and $n_\epsilon$ denotes the dimension of the state $\bm{\epsilon}$. Also, we re-parameterise the actual time index $t$ to a normalized $\bar{t}=\frac{t -t_0}{T}$, where $t_0$ is the starting time of each control loop, and $\bar{t} \in [0, 1]$.
$\bm{E}_0$ and $\bm{E}_{N_\epsilon}$ are equal to the beginning ($\bar{t}=0$) and end ($\bar{t}=1$) state configurations, respectively. 
Both the first and second derivatives of the B\'ezier curve are also B\'ezier curves -- the robot velocity and acceleration states can be derived analytically, \begin{equation}\label{differential}
\begin{aligned}
\dot{\bm{\epsilon}}(t) &= \bm{B}_{\dot{\epsilon}}(\bm{E}^{'}, \ \bar{t}) = \sum_{j=0}^{N_\epsilon-1} \tfrac{(N_\epsilon-1)!}{j!(N_\epsilon-1-j)!}\bar{t}^j(1-\bar{t})^{N_\epsilon-1-j}\bm{E}^{'}_{j}, \\
\ddot{\bm{\epsilon}}(t) &= \bm{B}_{\ddot{\epsilon}}(\bm{E}^{''}, \ \bar{t}) = \sum_{j=0}^{N_\epsilon-2} \tfrac{(N_\epsilon-2)!}{j!(N_\epsilon-2-j)!}\bar{t}^j(1-\bar{t})^{N_\epsilon-2-j} \bm{E}^{''}_j,     
\end{aligned}
\end{equation}
where $\bm{E}^{'}_j =\tfrac{N_{\epsilon}}{T_{\epsilon}}(\bm{E}_{j+1} - \bm{E}_j) $ and $\bm{E}^{''}_j = \tfrac{N_\epsilon(N_\epsilon-1)}{T^2_\epsilon}(\bm{E}_{j+2} - 2\bm{E}_{j+1} + \bm{E}_j)$.
The new control points $\bm{E}^{'}$ and $\bm{E}^{''}$ for $\dot{\bm{\epsilon}}(t)$ and $\ddot{\bm{\epsilon}}(t)$, respectively, can therefore be easily expressed as a function of $\bm{E}$. 
Then the discretized MPC in (\ref{discretized_MPC}) can be reformulated as follows, 
\begin{equation}
\begin{aligned}
\min_{\bm{E}} \  & L_E(\bm{E}), \\
s.t. \  & g_E(\bm{E} )\leq 0,
\end{aligned}
\end{equation}
where only the control points $\bm{E}$ serve as the decision variable which can express all the related position, velocity and acceleration states. 

For our high-DOF dual-arm mobile manipulator with hybrid position-velocity control, the incorporation of B\'ezier-curve parameterization in both the MPCs embodies the following advantages:
\begin{itemize}
\item No necessity for the definition of position/orientation, velocity, and acceleration as different decision variables since control points of the B\'ezier curves link them simultaneously. Consequently, equality constraints relating to approximate model state transitions at adjacent discretized time knots are eliminated. 
\item In the absence of inaccurate model state transition, whole-body MPC-W successfully generates consistent trajectories of $\bm{\epsilon}$ and $\dot{\bm{\epsilon}}$ for our position/velocity-controlled robot, leading to better tracking performance.
\item The time knot number $K$ can be much larger than the control point number $N_{\epsilon}$, reducing even more decision variables compared to discretized MPC, e.g. a straight line needs only two control points. 
\item Consequently, compared to the discretization approach, B\'ezier-curve-based MPC-T and MPC-W can run at much faster computation speeds by setting the same number of time knots and control points, and are not sensitive to horizon lengths.
\item The number of loss functions and constraints are reduced by exploiting control points to minimize and constrain velocities and accelerations, as well as the constraint number relating to obstacle avoidance.
\item Based the convex hull of B\'ezier curves, constraining the control points directly within a limit range $\bm{\epsilon}_{min}\leq \bm{E}_j\leq \bm{\epsilon}_{max},  \ j\in [0, \hdots, N_{\epsilon}]$ guarantees that the whole curve will satisfy the limit range in Fig. \ref{concept}(c). This property also applies to constraining velocity and acceleration limits using the corresponding control points $\bm{E}^{'}_j$ and $\bm{E}^{''}_j$. 
\item Predictive admittance control can also exploit the strengths of this efficient parametrization by representing the motion responses as B\'ezier curves.

\end{itemize}
Building on the bilevel-MPC framework and the B\'ezier-curve representation,  
we formulate the two MPCs are elucidated in Section \ref{subsec:stage1} and \ref{subsec:stage2}, respectively.  

\subsection{B\'ezier curve based MPC-T for Task Space Motion Planning} \label{subsec:stage1}
\subsubsection{End-effectors' translational motion}\label{position_bezier}
In our application scenario, we define the two end-effectors' translational motions $\bm{p}$ as B\'ezier curves with the formulation in (\ref{bezier_general}):
\begin{equation}\label{motion_planning_position}
	\begin{aligned}
		\bm{p}(t) &= \begin{bmatrix}\bm{p}^r(t) \\ \bm{p}^l(t)\end{bmatrix}= \bm{B}_{p}(\bm{P}, \ \bar{t}), 
	\end{aligned}
\end{equation}
where 
the superscripts $r$ and $l$ denote the $right$ and $left$ end-effector positions w.r.t. the inertial frame, respectively. $\bm{P} = [\bm{P}_0 \ \cdots \ \bm{P}_{N_p}]\in \mathbb{R}^{6\times N_p}$ is the set of control points, where $N_p+1$ denotes the number of control points.  
$\bar{t} = \frac{t-t_0}{T^{task}}$ is the scaling of the time index that depends on the time horizon $T^{task}$. At the beginning, the horizon $T^{task}$ is set with a pre-defined value. As the end-effectors approach the object target, $T^{task}$ decreases and is the remaining time period for two end-effectors to arrive at the given target. 

\subsubsection{End-effectors' rotational motion} \label{quaternion_bezier}
In this work, we employ quaternions to represent the end-effectors' rotational motions. To speed up the orientation motion planning, we first incorporate the B\'ezier-curve representation of two end-effectors' rotational motion into MPC-T. 
Typically, we can define the four components of a quaternion $\bm{\theta}$ trajectory to be B\'eizer curves. Equality unit-quaternion constraints can be added at each defined time knot. Since this setting cannot ensure all points on the curves are unit quaternions, the generated quaternion at specific time knots should be normalized after the control points are optimized, which are then sent to the whole-body MPC as rotational motion references. However, the equality constraints increase the burden on the MPC-T computation speed.
Another option is to apply more complex and high-nonlinear Spherical Linear Interpolation-based approaches \cite{shoemake1985animating, kim1995general, allmendinger2018coordinate}.
However, the equality constraints are still necessary conditions for trajectory optimization. 
To avoid the unit-quaternion constraints, the method in \cite{pu2020c2} builds a B-spline on logarithmic quaternions in $\mathbb{R}^3$ which are then mapped back to $SO(3)$. 
However, this method encounters difficulty when dealing with constant quaternions, which can occur when the robot adjusts whole-body motion while maintaining a constant orientation of the end-effectors.

To ensure that the entire orientation curve remains on a unit-quaternion manifold and to make the method applicable for constant and varying quaternions, we employ B\'ezier-curve parameterization on the unit-quaternion rotation formula which has the following expression:  
\begin{subequations}
\begin{align}
\bm{\theta} &= exp(\frac{\alpha}{2} \mathbf{u}) = \cos\frac{\alpha}{2} + \mathbf{u}\sin\frac{\alpha}{2}, \\
            &= \theta_w + \theta_x\bm{i} + \theta_y\bm{j} + \theta_z\bm{k},
\end{align}
\end{subequations}
where $\mathbf{u} = (u_x\bm{i}, u_y\bm{j}, u_z\bm{k})$ denotes a unit vector, which can be represented in a spherical coordinate system as $(\cos\beta \sin\gamma, \sin\beta \sin\gamma, \cos\gamma)$. $\alpha$ is the rotation angle around $\mathbf{u}$.
Consequently, the four components of a quaternion $(\theta_w, \theta_x, \theta_y, \theta_z)$ can be expressed depending on $(\alpha, \beta, \gamma)$:
\begin{subequations}
\begin{align}
    \theta_w &= \cos\frac{\alpha}{2}, \\
    \theta_x &= \sin\frac{\alpha}{2} \cos\beta \sin\gamma, \\
    \theta_y &= \sin\frac{\alpha}{2} \sin\beta \sin\gamma, \\
    \theta_z &= \sin\frac{\alpha}{2} \cos\gamma. 
\end{align}
\end{subequations}
We define $\bm{\psi}$ to combine two sets of $(\alpha, \beta, \gamma)$ for two end-effectors. Then $\bm{\psi}(t)$ can be represented as a B\'ezier curve, 
\begin{equation}\label{motion_planning_rotation}
\bm{\psi}(t) = \begin{bmatrix}\bm{\psi}^r(t) \\ \bm{\psi}^l(t)\end{bmatrix}= \bm{B}_{\psi}(\bm{\Psi}, \ \bar{t}),
\end{equation}
where $\bm{\psi}^r = [\alpha^r, \beta^r, \gamma^r]^T$ and $\bm{\psi}^l = [\alpha^l, \beta^l, \gamma^l]^T$. $\bm{\Psi} = [\bm{\Psi}_0 \ \cdots \ \bm{\Psi}_{N_\psi}]\in \mathbb{R}^{6\times N_\psi}$. In the rest of the paper, we use $\bm{\theta} = [(\bm{\theta}^r)^T, \ (\bm{\theta}^l)^T]^T$ to denote two end-effectors' quaternion. We define $\bm{\theta}(t) = \bm{f}_{\psi}^{\theta}(\bm{\psi}(t))$ to transform the trajectory $\bm{\psi}(t)$ to the quaternion trajectory $\bm{\theta}(t)$. 
To enable zero quaternion velocity and acceleration, we can easily regulate equality constraints of the last two and three control points of $\bm{\Psi}$. 
Compared to direct B\'ezier-curve parameterization on the four components of a quaternion, our method eliminates the equality unit-quaternion constraints and ensures that all points on the quaternion curve belong to $SO(3)$.
This facilitates fast execution of MPC-T, as detailed in the comparative simulation results in Section \ref{mpce_comparision_efficiency}.

\begin{figure}[t]
\centering
\includegraphics[height=4.cm]{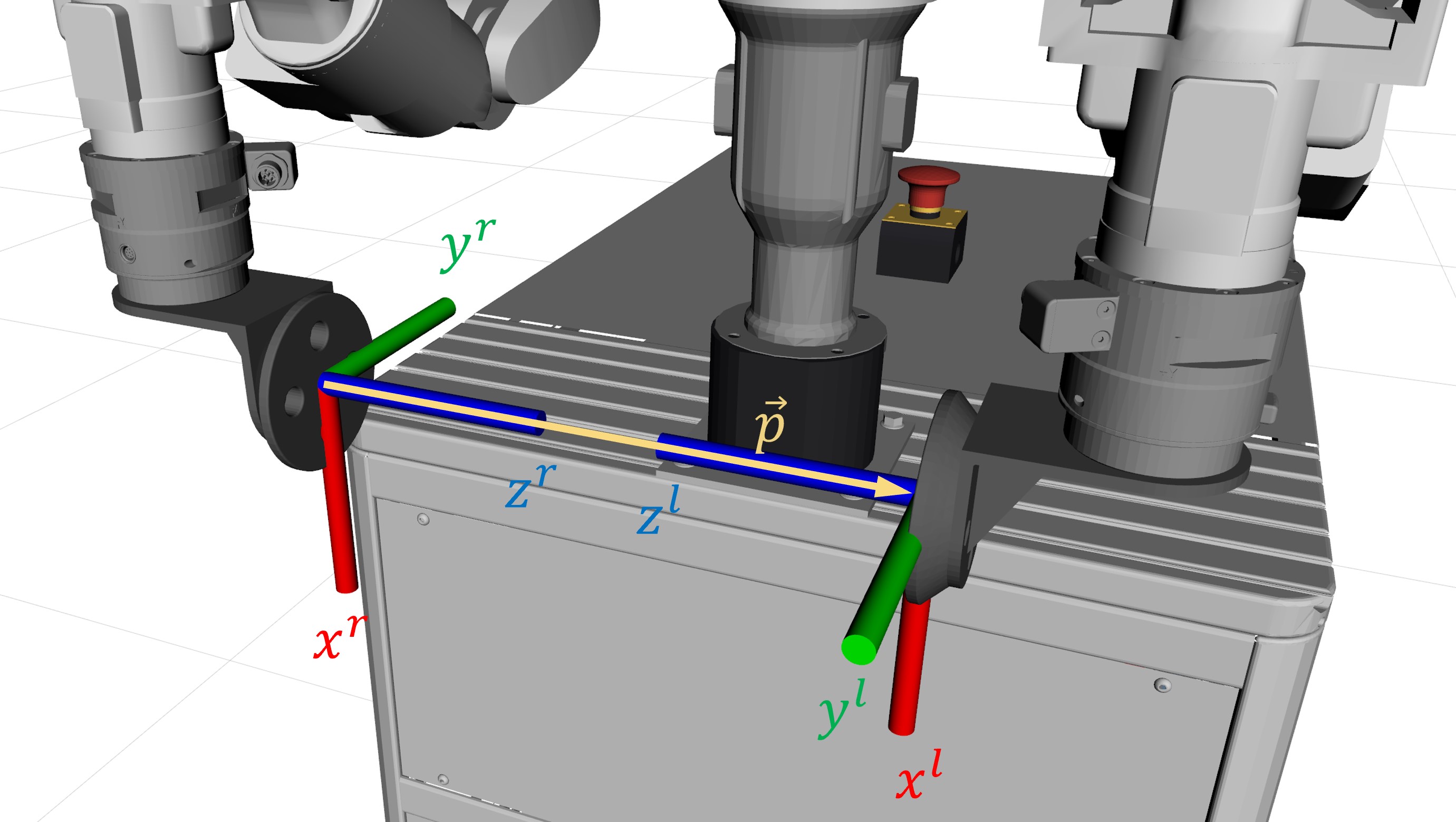}
\caption{End-effector frames when two palms are set to be parallel. The red, green, and blue lines denote the $x$, $y$, and $z$ axes, respectively.
}
\label{end_effector_frames}
\end{figure}

\subsubsection{MPC-T formulation}\label{formulation_MPC_T}
Based on the B\'ezier-curve representation of the two end-effectors' translational and rotational motion ($\bm{p}(t), \bm{\psi}(t)$), we build MPC-T depending on the corresponding control points ($\bm{P}, \bm{\Psi}$):
\begin{equation}\label{mpc_e_bezier}
\begin{aligned}
\min_{\bm{P}, \bm{\Psi}} \  & L_{P, \Psi}(\bm{P}, \bm{\Psi}),\\
s.t. \ & g_{P, \Psi}(\bm{P}, \bm{\Psi} )\leq 0, 
\end{aligned}
\end{equation}
where $L_{P, \Psi}(\bm{P}, \bm{\Psi})$ is the cost function consisting of two summed terms, including $L^1_{P, \Psi}(\bm{P}, \bm{\Psi})$ that addresses the relative pose between the two end-effectors and $L^2_{P, \Psi}(\bm{P}, \bm{\Psi})$ that minimizes trajectory length, velocity and acceleration. 
$g_{P, \Psi}(\bm{P}, \bm{\Psi} )$ refers to the constraints accounting for trajectory continuity, obstacle avoidance, and base motion capabilities. 

Next, we look at the components of the MPC-T cost functions in more detail. 

\noindent (a). When the robot is transporting an object (e.g. a box in this paper), the two collaborative end-effectors should maintain constant relative motion w.r.t. the object. However, this relationship is not always satisfied since the orientations of the end-effectors evolve throughout the entire process from the current states to the goal states. Therefore, we set two palms to face each other as aligned as possible, and the corresponding cost function is formulated at $K^{task}+1$ time knots as: 
\begin{equation}\label{cost_ob_task}
\begin{aligned}
 L^1_{P, \Psi}(\bm{P}, \bm{\Psi}) &  \\
= \sum_{i=0}^{K^{task}} & \lVert [\underbrace{\bm{f}_\theta^{R}(\bm{\theta}^r_{i})\hat{\bm{x}}}_{\bm{x}^r_i}]^T \vec{\bm{p}}_i\rVert_{\bm{w}_{x}}^2 + \lVert [\underbrace{\bm{f}_\theta^{R}(\bm{\theta}^r_{i})\hat{\bm{y}}}_{\bm{y}^r_i}]^T  \vec{\bm{p}}_i\rVert_{\bm{w}_{y}}^2 +  \\
&\lVert [\underbrace{\bm{f}_\theta^{R}(\bm{\theta}^l_{i})\hat{\bm{x}}}_{\bm{x}^l_i}]^T \vec{\bm{p}}_i\rVert_{\bm{w}_{x}}^2 + \lVert [\underbrace{\bm{f}_\theta^{R}(\bm{\theta}^l_{i})\hat{\bm{y}}}_{\bm{y}^l_i}]^T  \vec{\bm{p}}_i\rVert_{\bm{w}_{y}}^2,  
\end{aligned}
\end{equation}
where the subscripts using ``$\bm{w}$" represent the weight matrices, $\hat{\bm{x}} = [1 \ 0 \ 0]^T$ and $\hat{\bm{y}} = [0 \ 1 \ 0]^T$ and $\vec{\bm{p}_i} = \bm{p}^l_i - \bm{p}^r_i$ represents the vector along two-palm centers, as shown in Fig. \ref{end_effector_frames}. 
 $\bm{f}_\theta^R(\cdot)$ is a function that transforms a quaternion to a rotation matrix. 
$(\bm{x}^r, \bm{y}^r)$ and $(\bm{x}^l, \bm{y}^l)$ denote the $x$ and $y$ axes of two palm frames. The corresponding cost function (\ref{cost_ob_task}) facilitates $\vec{\bm{p}}$ to be as perpendicular as possible to $\bm{x}^r$, $\bm{y}^r$, $\bm{x}^l$, and $\bm{y}^l$. Based on the B\'ezier-curve representation of position and quaternion trajectories in Section \ref{position_bezier} and \ref{quaternion_bezier}, all the variables in this Section \ref{formulation_MPC_T} can be expressed in terms of the corresponding control points which are the decision variables.  

\noindent (b). According to the property of B\'ezier curves, the minimization of the deviation of adjacent control points leads to the minimization of the curve length. The direct minimization of each control point value results in the minimization of each point value on the curve. We define the following cost functions to minimize the lengths of the curves $\bm{p}(t), \bm{\psi}(t), \dot{\bm{p}}(t), \dot{\bm{\psi}}(t)$, and minimize the velocity and acceleration values on the derivative curves $\dot{\bm{p}}(t), \dot{\bm{\psi}}(t), \ddot{\bm{p}}(t), \ddot{\bm{\psi}}(t)$:
\begin{equation}\label{cost_ob2}
	\begin{aligned}
		 L^2_{P, \Psi}(\bm{P}, \bm{\Psi}) = \sum_{j=0}^{N_{p}-1} & \lVert \bm{P}^{'}_{j} \rVert_{\bm{w}_{\dot{p}}}^2+  \sum_{j=0}^{N_{\psi}-1} \lVert \bm{\Psi}^{'}_{j}  \rVert_{\bm{w}_{\dot{\psi}}}^2+ \\
  		\sum_{j=0}^{N_{p}-2} & \lVert \bm{P}^{''}_{j} \rVert_{\bm{w}_{\ddot{p}}}^2+ \sum_{j=0}^{N_{\psi}-2}  \lVert \bm{\Psi}^{''}_{j} \rVert_{\bm{w}_{\ddot{\psi}}}^2,  
	\end{aligned}
\end{equation}
where $\bm{P}^{'}_j$, $\bm{P}^{''}_j$, $\bm{\Psi}^{'}_j$, $\bm{\Psi}^{''}_j$ denote the $j$-th control points of the B\'ezier curves $\dot{\bm{p}}(t)$, $\ddot{\bm{p}}(t)$, $\dot{\bm{\psi}}(t)$ and $\ddot{\bm{\psi}}(t)$. 
Based on (\ref{differential}), they are expressed analytically as: $\dot{\bm{p}}(t) = \bm{B}_{\dot{p}}(\bm{P}^{'}, \bar{t})$, $\ddot{\bm{p}}(t) = \bm{B}_{\ddot{p}}(\bm{P}^{''}, \bar{t})$, $\dot{\bm{\psi}}(t) =\bm{B}_{\dot{\psi}}(\bm{\Psi}^{'}, \bar{t})$,  $\ddot{\bm{\psi}}(t) = \bm{B}_{\ddot{\psi}}(\bm{\Psi}^{''}, \bar{t})$.
With the B\'ezier-curve representation, utilization of the few control points in (\ref{cost_ob2}) can shape the whole trajectories. 

Next we turn our attention to the terms that represent the MPC-T constraints.

\noindent (a). With B\'ezier curve parameterization, initial and terminal constraints can be expressed using the control points:
\begin{subequations}\label{constraint_ob_constraint}
\begin{align}
& \bm{p}_0 =\bm{P}_0 = \bm{p}_{act}, \
\bm{p}_{K^{task}} =\bm{P}_{N_p} = \bm{p}_{goal}, \label{equality1}\\
&\dot{\bm{p}}_{K^{task}} =\tfrac{N_p}{T^{task}}(\bm{P}_{N_p} - \bm{P}_{N_p-1})= \bm{0},   \label{equality3}\\
& \ddot{\bm{p}}_{K^{task}} =\tfrac{N_p(N_p-1)}{T^2_{plan}}(\bm{P}_{N_p} - 2\bm{P}_{N_p-1} +\bm{P}_{N_p-2})= \bm{0},  \label{equality4} \\
& \bm{\theta}_0 =\bm{\Theta}_0 = \bm{\theta}_{act}, \
\bm{\theta}_{K^{task}} = \bm{\theta}_{goal}, \label{equality6}\\
&\dot{\bm{\psi}}_{K^{task}}  = \tfrac{N_{\psi}}{T^{task}} (\bm{\Psi}_{N_{\psi}} - \bm{\Psi}_{N_{\psi}-1}) = \bm{0}, \label{equality7} \\
&\ddot{\bm{\psi}}_{K^{task}} =\tfrac{N_\psi(N_\psi-1)(\bm{\Psi}_{N_\psi} - 2\bm{\Psi}_{N_\psi-1} +\bm{\Psi}_{N_\psi-2})}{(T^{task})^2} =  \bm{0}, \label{equality8} 
\end{align}
\end{subequations}
where 
the first and last control points are used directly to constrain the position and orientation states, shown in (\ref{equality1}) and (\ref{equality6}). In (\ref{equality3}) and (\ref{equality4}), the last two and three control points are used to enforce the zero translational velocity and acceleration at the target, respectively. 
We apply the constraints on $\bm{\Psi}$ in (\ref{equality7}) and (\ref{equality8}) to enforce $\dot{\bm{\theta}} = \bm{0}$ and $\ddot{\bm{\theta}}=\bm{0}$ at the terminal timing result. With the B\'ezier-curve representation, velocity and acceleration are computed analytically, which avoids inaccurate model-state transition due to approximations. 

\noindent (b). To enable the first stage to generate collision-free end-effectors' trajectories which serve as motion references for the second stage, a safety constraint is defined as follows, 
\begin{equation}\label{inequality1}
\lVert \bm{p}^m_i - \bm{p}^{obstacle}_k\rVert \geq d_{safe}, \forall i\in[0, K^{task}], 
\end{equation}
where $d_{safe}$ denotes a safe distance. $\bm{p}^{m} = \frac{\bm{p}^r + \bm{p}^l}{2}$ represents the middle point position of two end-effectors and $\bm{p}^{obstacle}_k$ is the $k$-th obstacle position. 

\noindent (c). The end-effectors' motion in MPC-T should also account for the mobile base capability.
Considering that the mobile base can move significantly faster than the end-effectors, we define the following constraints to enable the end-effectors' motion to satisfy the base velocity and acceleration limits, 
\begin{subequations}\label{constraint_ob_constraint2}
\begin{align}
& \dot{\bm{p}}_{min} \leq \bm{P}^{'}_{j} \leq \dot{\bm{p}}_{max}, \forall j\in[0, N_p-1]  \label{inequality2}\\
& \ddot{\bm{p}}_{min} \leq \bm{P}^{''}_{j} \leq \ddot{\bm{p}}_{max}, \forall j\in[0, N_p-2], \label{inequality3}
\end{align}
\end{subequations}
where 
$(\dot{\bm{p}}_{min}, \dot{\bm{p}}_{max})$ and $(\ddot{\bm{p}}_{min}, \ddot{\bm{p}}_{max})$ define the base velocity and acceleration limits. 
The constraints on $\bm{P}^{'}_j$ and $\bm{P}^{''}_j$ within a limit range are beneficial and enable all points on the velocity and acceleration curves to be strictly constrained \cite{fernbach2020c}, resulting in a reduction of the constraint number while maintaining low constraint nonlinearity. 

Based on the cost functions in (\ref{cost_ob_task})(\ref{cost_ob2}) and constraints in (\ref{constraint_ob_constraint})(\ref{inequality1})(\ref{constraint_ob_constraint2}), the first-stage MPC-T can be solved to generate the optimized control points of the corresponding B\'ezier curves. In this MPC-T, the control point number is set to 8, allowing for the representation of complex trajectories. In addition to accurate model-state transition due to B\'ezier curve parameterization, we set $K^{task}$ to be a constant and relatively small value for two additional reasons. Due to the adoption of a two-stage MPC structure, the second stage is capable of refining the number of time nodes in a relatively short time span. Moreover, when the robot is far from the goal, the end-effector trajectory can be sparse. 
As the robot approaches the goal and the time horizon decreases, the generated trajectory becomes more accurate. 
Then we transform the generated control points to achieve the end-effectors' trajectories which serve as the motion references of the second MPC-W, 
\begin{subequations}\label{consistent_ref}
\begin{align}
\bm{p}^{ref} &= \bm{B}_{p}(\bm{P}, \ \bar{t}=\tfrac{t-t_0}{T^{task}}), \quad \quad \ \text{if} \ t-t_0 \leq T^{task}, \\
& = \bm{B}_{p}(\bm{P}, \ \bar{t}=1), \quad \quad \quad \ \ \ \text{others},\\
\bm{\theta}^{ref} &= \bm{f}_{\psi}^{\theta}(\bm{B}_{\psi}(\bm{\Psi}, \ \bar{t}=\tfrac{t-t_0}{T^{task}})), \ \text{if} \ t-t_0 \leq T^{task}, \\
& =  \bm{f}_{\psi}^{\theta}(\bm{B}_{\psi}(\bm{\Psi}, \ \bar{t}=1)), \quad \ \ \ \text{others}.
\end{align}
\end{subequations}
Since the horizons of two-stage MPCs are different, the above processing ensures that the terminal-state references of MPC-T are tracked by MPC-W. 

\subsection{B\'ezier curve based MPC-W for Compliant Whole-body Motion Generation}\label{subsec:stage2}

\subsubsection{B\'ezier curve parameterization for joint configuration}

The dual-arm mobile manipulator named EVA has a mobile base with 3 DOFs, and a humanoid upper body with 15 DOFs.
The mobile base with four mecanum wheels is velocity-controlled and the upper body joints are controlled using position signals. A simple rigid end-effector is installed at the end of each arm and the wrists are each equipped with a force sensor.
The joint configuration is denoted as follows, 
\begin{equation}
	\bm{q}^T = [\bm{q}^{base} \ \ \bm{q}^{chest} \ \ \bm{q}^{head} \  \ \bm{q}^{arm} ],
\end{equation}
where $\bm{q}\in \mathbb{R}^{n_{dof}=18}$ and $n_{dof}$ denotes the robot DOF. The base can only operate on flat terrain, and $\bm{q}^{base} \in \mathbb{R}^3$ includes the 2-DOF prismatic motion and 1-DOF turning motion of the mobile base. $\bm{q}^{chest}\in \mathbb{R}^{1}$, $\bm{q}^{head}\in \mathbb{R}^{2}$, and $\bm{q}^{arm}\in \mathbb{R}^{12}$ represents the chest, head, and dual-arm joint configurations. 

Based on the efficient B\'ezier curve representation, we define the joint trajectory as follows, 
\begin{equation}\label{bezier}
	\bm{q}(t) = \bm{B}_q(\bm{Q}, \ \bar{t}), 
\end{equation}
where $\bm{Q} = [\bm{Q}_0 \ \cdots \ \bm{Q}_{N_q}]\in \mathbb{R}^{n_{dof}\times (N_q +1)}$ denotes the control points. 
(\ref{bezier}) will be integrated into MPC-W, and $\bm{Q}$ is optimized to generate whole-body motion. With this continuous representation, the generated position trajectories of the upper-body joints and velocity trajectories of the mobile base are consistent. 

\subsubsection{B\'ezier Curve based Admittance Control}
Since our dual-arm mobile manipulator is position/velocity-controlled 
and the force sensors are only available at two end-effectors, admittance control can be applied to enable compliance at two end-effectors' operational space, by transferring force deviation to end-effectors' motion responses. 

To track the interaction force reference $\bm{F}^{ref}$, based on operational-space dynamics and feedback control, the optimized interaction force at two end-effectors $\bm{F}^{opt}$ can be built depending on the actual interaction force $\bm{F}^{act}$ and the deviation between the task-space motion reference $\bm{p}^{ref}$ and the optimized motion $\bm{p}^{opt}$: 
\begin{equation}\label{F_ext_controller}
	\begin{aligned}
		\bm{F}^{opt} &= \bm{F}^{act} + \bm{\Lambda}\ddot{\tilde{\bm{p}}} + \bm{K}\tilde{\bm{p}} + \bm{D}\dot{\tilde{\bm{p}}},
	\end{aligned}
\end{equation}
where $\tilde{\bm{p}} = \bm{p}^{opt} - \bm{p}^{ref}$ denotes the motion response. Although this formulation resembles the formulation of an impedance controller, we introduce it as a constraint in MPC-W. $\bm{\Lambda}\in \mathbb{R}^{6\times 6}$ denotes the corresponding inertia. $\bm{K}\in \mathbb{R}^{6\times 6}$ and $\bm{D}\in \mathbb{R}^{6\times 6}$ denote the stiffness and damping matrices. The motion response $\tilde{\bm{p}}(t)$ is optimized to update the motion reference and minimize the error between $\bm{F}^{opt}$ and the force reference $\bm{F}^{ref}$ by satisfying whole-body constraints, therefore, this works as an admittance controller. Given that the mobile base can only reach a maximum acceleration of $1m/s^2$, and considering the small mass $0.33kg$ of each end-effector, $\bm{\Lambda}\ddot{\tilde{\bm{p}}}$ can be ignored. 
With the B\'ezier curve parameterization, the motion response $\tilde{\bm{p}}$ is also built as:
\begin{equation}
	\tilde{\bm{p}}(t) = \bm{B}_{\tilde{p}}(\tilde{\bm{P}}, \ \bar{t}),
\end{equation}
where $\tilde{\bm{P}} = [\tilde{\bm{P}}_0 \ \cdots \ \tilde{\bm{P}}_{N_{\tilde{p}}}] \in \mathbb{R}^{6\times (N_{\tilde{p}} +1)}$ denotes the control points of the motion response. Since $\tilde{\bm{p}}$, $\dot{\tilde{\bm{p}}}$ and $\ddot{\tilde{\bm{p}}}$ share the same decision variables $\tilde{\bm{P}}$, 
only $\tilde{\bm{P}}$ is optimized to track the force reference while updating the motion reference.  

\subsubsection{MPC-W formulation}
To optimize the control points of the joint and motion-response trajectories, MPC-W is built: 
\begin{equation}\label{mpc_w_bezier}
\begin{aligned}
\min_{\bm{Q}, \tilde{\bm{P}}} \  & L_{Q, \tilde{P}}(\bm{Q}, \tilde{\bm{P}}), \\
s.t. \ & g_{Q, \tilde{P}}(\bm{Q}, \tilde{\bm{P}})\leq 0, 
\end{aligned}
\end{equation}
where $L_{Q, \tilde{P}}(\bm{Q}, \tilde{\bm{P}})$ denotes the MPC-W cost function that includes $L^1_{Q, \tilde{P}}(\bm{Q}, \tilde{\bm{P}})$ for motion and force tracking, $L^2_{Q, \tilde{P}}(\bm{Q}, \tilde{\bm{P}})$ for upper-body motion minimization, and $L^3_{Q, \tilde{P}}(\bm{Q}, \tilde{\bm{P}})$ for minimizing velocities and accelerations of joint and motion-response trajectories.  
$g_{Q, \tilde{P}}(\bm{Q}, \tilde{\bm{P}})$ represents the MPC-W constraints accounting for the whole-body trajectory continuity, joint limits, admittance controller, and obstacle avoidance. 

First, the MPC-W cost functions are formulated as follows.  

\noindent (a). The first cost function relating to motion and force tracking is built at $K^{wb}+1$ time knots:
\begin{subequations}\label{cost_ad}
\begin{align}
L^1_{Q, \tilde{P}}(\bm{Q}, \tilde{\bm{P}}) = \sum_{i=0}^{K^{wb}} &\lVert\bm{f}^{p}_q(\bm{q}_i) - \bm{p}^{ref}_i - \bm{f}^R_{q}(\bm{q}_i)\tilde{\bm{p}}_i\rVert_{\bm{w}_{p}}^2 + \label{cost_ad1}\\
&\lVert\bm{f}^{\theta}_q(\bm{q}_i) -  \bm{\theta}^{ref}_i)\rVert_{\bm{w}_{\theta}}^2 +  \label{cost_ad2} \\
&\lVert \bm{F}^{opt}_i - \bm{F}^{ref}_i \rVert_{\bm{w}_{f}}^2,   \label{cost_ad3}
\end{align}
\end{subequations}
where $\bm{f}^{p}_q(\cdot)$, $\bm{f}^R_q(\cdot)$ and $\bm{f}^{\theta}_q(\cdot)$ denote the functions that transform the joint configuration to two end-effector positions, rotation matrices and quaternions, respectively. 
(\ref{cost_ad1}) and (\ref{cost_ad2}) relate to tracking the position and quaternion references, respectively. Based on the admittance controller in (\ref{F_ext_controller}), (\ref{cost_ad3}) is used as force tracking which optimizes $\tilde{\bm{p}}$ to update the translational motion reference in (\ref{cost_ad1}). For dual-arm mobile manipulation, the force reference $\bm{F}^{ref}$ is more intuitive to be provided in the local frames of two end-effectors which is designed to be a 5-order curve. Consequently, we build the admittance control in the same frames. In this way, (\ref{cost_ad}) harmonizes the motion and force tracking.

\noindent (b). Since the mobile base has a larger range of motion compared to the dual arms, we define the second cost function to ensure minimal arm-joint motion while maintaining feasibility and future manipulability:
\begin{equation}\label{cost_ad5} 
L^2_{Q, \tilde{P}}(\bm{Q}, \tilde{\bm{P}}) = \sum_{j=0}^{N_{q}} \lVert \bm{Q}^{u}_{j} \rVert_{\bm{w}_{u}}^2,  
\end{equation}
where $\bm{Q}^u \in \mathbb{R}^{15\times N_q}$ represents the control points of upper-body joint-position B\'ezier curves and equals to the last 15 rows of $\bm{Q}$ in (\ref{bezier}).

\noindent (c). Similar to (\ref{cost_ob2}), the following cost functions are defined to minimize the velocities and accelerations of points on the joint-motion trajectory and motion-response trajectory using the corresponding control points:
\begin{equation}\label{cost_ad_mpcw}
\begin{aligned}
L^3_{Q, \tilde{P}}(\bm{Q}, \tilde{\bm{P}}) = \sum_{j=0}^{N_{\tilde{p}}-1} & \lVert \tilde{\bm{P}}^{'}_{j} \rVert_{\bm{w}_{\dot{\tilde{p}}}}^2 +  \sum_{j=0}^{N_q-1} \lVert \bm{Q}^{'}_{j}  \rVert_{\bm{w}_{\dot{q}}}^2 +    \\
  \sum_{j=0}^{N_{\tilde{p}}-2} & \lVert \tilde{\bm{P}}^{''}_{j} \rVert_{\bm{w}_{\ddot{\tilde{p}}}}^2 +  \sum_{j=0}^{N_q-2} \lVert \bm{Q}^{''}_{j}  \rVert_{\bm{w}_{\ddot{q}}}^2,  
\end{aligned}
\end{equation}
where $\tilde{\bm{P}}^{'}_j$, $\tilde{\bm{P}}^{''}_j$, $\bm{Q}^{'}_j$ and $\bm{Q}^{''}_j$ denote the $j$'s control point of $\dot{\bm{p}}(t)$, $\ddot{\bm{p}}(t)$, $\dot{\bm{q}}(t)$, $\ddot{\bm{q}}(t)$, respectively, which also embody the analytical expressions using (\ref{differential}): $\dot{\bm{q}}(t) = \bm{B}_{\dot{q}}(\bm{Q}^{'}, \bar{t})$, $\ddot{\bm{q}}(t) = \bm{B}_{\ddot{q}}(\bm{Q}^{''}, \bar{t})$, $\dot{\tilde{\bm{p}}}(t) = \bm{B}_{\dot{\tilde{p}}}(\tilde{\bm{P}}^{'}, \bar{t})$, $\ddot{\tilde{\bm{p}}}(t) = \bm{B}_{\ddot{\tilde{p}}}(\tilde{\bm{P}}^{''}, \bar{t})$.

Next, we address the MPC-W constraints.

\noindent (a). Based on the B\'ezier curve parameterization, the initial conditions of the whole-body motion, admittance control model, and joint limits can be expressed using the control points. The constraints are listed as follows,
\begin{subequations}\label{constraint_ad}
\begin{align} 
&\bm{F}^{ext}_0 =\bm{F}^{ext}_{act} \ \text{and} \ \tilde{\bm{P}}_0 = \bm{0},  \label{mpcw_1}\\
&\bm{F}^{opt}_i = \bm{F}^{act} + \bm{K}\tilde{\bm{p}}_i + \bm{D}\dot{\tilde{\bm{p}}}_i,  \forall i\in[0, K^{wb}], \label{admittance_model}\\  
&\bm{q}_0 = \bm{Q}_0  = \bm{q}_{act}, \label{mpcw_3}\\
&\bm{q}_{min}\leq \bm{Q}_j \leq \bm{q}_{max}, \forall j\in[0, N_{q}], \label{mpcw_5}\\
&\dot{\bm{q}}_{min}\leq \tfrac{N_q(\bm{Q}_{j+1}-\bm{Q}_j) }{T^{wb}}\leq \dot{\bm{q}}_{max}, \forall j\in[0, N_{q} - 1],\label{mpcw_6}
\end{align}
\end{subequations}
where the constraint in (\ref{mpcw_1}) relates to the initial state of interaction force, and (\ref{admittance_model}) represents the admittance control model. (\ref{mpcw_3}) corresponds to the initial joint state. (\ref{mpcw_5}) and (\ref{mpcw_6}) apply control points of the joint trajectory to ensure that all points on the joint trajectory satisfy the joint limits.

\noindent (b). Although obstacle avoidance of end-effectors is already integrated in the first-stage MPC-T, we should also add this constraint in this second-stage MPC-W to ensure strict safety requirements. Furthermore, MPC-W should also satisfy the collision avoidance constraint between the mobile base and the dynamic obstacle. 
Instead of covering the whole robot as a big sphere, we set safety constraints individually for the end-effectors and the mobile base to enable the robot to navigate narrow paths, shown in Section \ref{experiment2}. The safety constraints are listed as follows, 
\begin{subequations}\label{constraint_ad2} 
\begin{align} 
&\lVert \tfrac{\bm{f}^{p^r}_q(\bm{q}_i) + \bm{f}^{p^l}_q(\bm{q}_i)}{2} - \bm{p}^{obstacle}_k\rVert \geq d_{safe},  \forall i\in[0, K^{wb}] \label{mpcw_7}\\
&\lVert \bm{Q}^{base}_j - \bm{p}^{obstacle}_k\rVert \geq d_{safe},  \forall j\in[0, N_{q}] \label{base_obstacle_avoidance}
\end{align}
\end{subequations}
where (\ref{mpcw_7}) denotes the safety constraint between the middle point of two end-effectors and the obstacle. $\bm{f}^{p^r}_q(\cdot)$ and $\bm{f}^{p^l}_q(\cdot)$ transform the joint configuration to the right and left end-effector positions, respectively. (\ref{base_obstacle_avoidance}) is another safety constraint between the mobile base and the obstacle. 
Due to the convex-hull property of B\'ezier curves, the obstacle avoidance relating to the base is built using the control points $\bm{Q}^{base}$ which are the first two rows of $\bm{Q}$ and denote the base translational motion. Since the end-effectors are located at the end of the robot tree structure, their motions in MPC-W embody high non-linearity depending on all joint motions. With the help of MPC-T, the generated collision-free end-effectors' motion in MPC-T lowers the burden of MPC-W which can handle the end-effectors' obstacle avoidance efficiently. 
In addition, since the base motion typically has low non-linearity, the constraint relating to base obstacle avoidance only increases the complexity of MPC-W to a small extent. 

Based on the cost functions in (\ref{cost_ad}), (\ref{cost_ad5}) and (\ref{cost_ad_mpcw}), and the constraints in (\ref{constraint_ad}), MPC-W is solved online to generate the whole-body joint trajectories. 
Based on the B\'ezier curve parameterization on joint motion trajectory, the consistent position and velocity commands for the upper-body joints and the mobile base can be achieved: 
\begin{subequations}
\begin{align}
    \bm{S}^{u}\bm{q}(t_0 + t_{loop}) &= \bm{S}^{u}\bm{B}_q(\bm{Q}, \bar{t}=\bm{S}^{u}\tfrac{t_{loop}}{T^{wb}})\\
 \bm{S}^{base}\dot{\bm{q}}(t_0 + t_{loop}) &=\bm{S}^{base}\bm{B}_{\dot{q}}(\bm{Q}^{'}, \bar{t}=\tfrac{t_{loop}}{T^{wb}})
\end{align}
\end{subequations}
where $t_{loop}$ is the period of a control loop which is equal to $0.02s$ in this paper. $\bm{S}^{u}$ and $\bm{S}^{base}$ are used to select the corresponding upper-body and mobile-base components. Since the mobile-base motion is built w.r.t. the inertial frame, the commands for the mobile base should further be transformed to the base's local frame.

\begin{figure}[t]
	\centering
	\includegraphics[height=5cm]{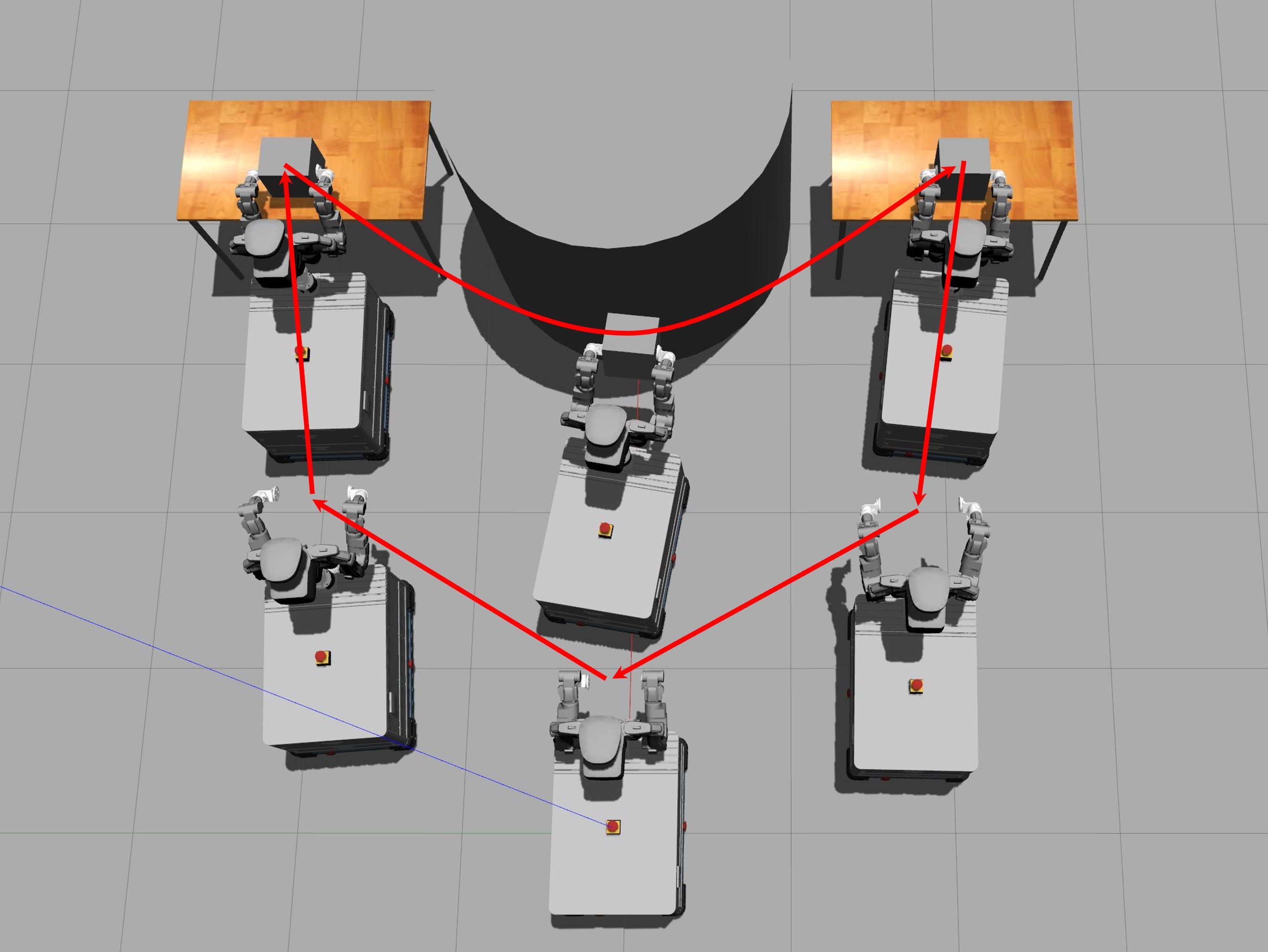}
	\caption{Simulation scenario that involves a static obstacle and specifies the directions in which the robot moves. 
	}
	\label{simulation_scenario}
\end{figure}

\begin{table*}[t]
\caption{Computation time comparison of MPC-T using three methods}
\label{comparison_mpcE}
\begin{tabular}{p{4.6cm}p{4.4cm}p{3.5cm}p{3.9cm}}
\hline\noalign{\smallskip}
Items & Discretization on $\bm{p}$, $\bm{\theta}$ &  B\'ezier curve of $\bm{p}$, $\bm{\theta}$ & B\'ezier curve of $\bm{p}$, $\bm{\psi}$ (ours)  \\
\noalign{\smallskip}\hline\noalign{\smallskip}
Time knot number & $K^{task}+1=8$& $K^{task}+1=8$& $K^{task}+1=8$ \\
Model dimension & $14$ & $14$ & $12$\\
Control point number& - & $N+1 = 8$ & $N+1 = 8$ \\
Decision variables & $(\bm{p}, \dot{\bm{p}}, \bm{\theta}, \dot{\bm{\theta}})_{[0 \cdots K^{task}]}$ & $(\bm{P}, \bm{\Theta})_{[0 \cdots N]}$ &  $(\bm{P}, \bm{\Psi})_{[0 \cdots N]}$ \\
Decision variable number& $2(K^{task}+1) \times 14 = 224$ & $ (N+1) \times 14 = 112$&  $(N+1) \times 12 = 96$\\
Unit-quaternion constraint & Yes & Yes & No\\
Unit-quaternion guarantee on all points & No & No & Yes \\
Average MPC-T computation time & $60.8ms$& $19.4ms$& $6.2ms$ \\
Standard deviation & $36.8ms$& $8.9ms$& $2.4ms$  \\
\noalign{\smallskip}\hline\noalign{\smallskip}
\end{tabular}
\end{table*}

\begin{table*}[ht]
\caption{Comparison for motion tracking performance between discretization and our B\'ezier-curve-based whole-body MPC-W, without obstacle avoidance or force control}
\label{comparison}       
\begin{tabular}{p{3.6cm}p{3.2cm}p{3.2cm}p{3cm}p{3cm}}
\hline\noalign{\smallskip}
Items & \multicolumn{2}{c}{Discretized MPC-W} & \multicolumn{2}{c}{B\'ezier-curve-based MPC-W (ours)} \\
\noalign{\smallskip}\hline\noalign{\smallskip}
Horizon& 5s& 5s& 5s&5s \\
Time knot number& $K^{wb} +1=6$& $K^{wb} +1=26$& $K^{wb}+1=6$& $K^{wb}+1=26$ \\
Control point number& - & - & $N_q+1=6$& $N_q+1=6$ \\
Decision variable & $(\bm{q}, \dot{\bm{q}})_{[0 \cdots K^{wb}]}$& $(\bm{q}, \dot{\bm{q}})_{[0 \cdots K^{wb}]}$& $\bm{Q}_{[0 \cdots N_q]}$& $\bm{Q}_{[0 \cdots N_q]}$ \\
Decision variable number & $2(K^{wb} +1)\times n_{dof} = 216$ & $2(K^{wb} +1)\times n_{dof} = 936$&  $(N_q+1) \times n_{dof} = 108$& $(N_q+1) \times n_{dof} = 108$\\
Average computation time & $12ms$& $60.1ms$& $9ms$ &$12.5ms$ \\
Standard deviation & $5ms$& $35.7ms$& $3.5ms$ &$3ms$ \\
Mean absolute tracking error  & 21.6\% & 6.3\% & 7.7\% & 5.1\%\\
\noalign{\smallskip}\hline\noalign{\smallskip}
\end{tabular}
\end{table*}

\section{Simulations and Experiments}\label{demo}

In the simulations and experiments, we utilize OpTaS \cite{mower2023optas}, a task specification Python library, and Casadi \cite{Andersson2019}, to construct the bilevel MPCs, and use KNITRO as the nonlinear problem solver \cite{byrd2006k}. 
The calculations are conducted on a workstation equipped with an Intel i9 CPU.

To show the performance of the bilevel MPCs, we conduct several comparative simulations in ROS-GAZEBO between the discretized method and our B\'ezier-curve-based approach. These simulations encompass several aspects, including the computation efficiency of MPC-T (Section \ref{mpce_comparision_efficiency}), tracking performance of MPC-W (Section \ref{tracking_performance}), computation performance of MPC-W under different robot capabilities and horizons (Section \ref{capabilities}). Additionally, we analyze the impact of MPC-T on MPC-W in Section \ref{mpce_comparision}. 

To validate the efficiency, robustness, and capabilities of our bilevel-MPC framework, our experiments focus on several key scenarios. The first experiment relates to dynamic obstacle avoidance and push recovery, demonstrating the effective guidance of MPC-T on MPC-W and compliant force control (Section \ref{experiment3}). We explore the robot's navigation capability in narrow paths to show how MPC-W corrects trajectories generated by MPC-T in Section \ref{experiment2}. We also examine coordinated motions of the lower and upper bodies to illustrate consistent whole-body motion in Section \ref{experiment1}. Furthermore, manipulation with object-shape adaptation is investigated to highlight the robustness of our bilevel-MPC framework (Section \ref{experiment4}).

\begin{figure}[t]
	\centering
 	\includegraphics[width = 0.94\linewidth]{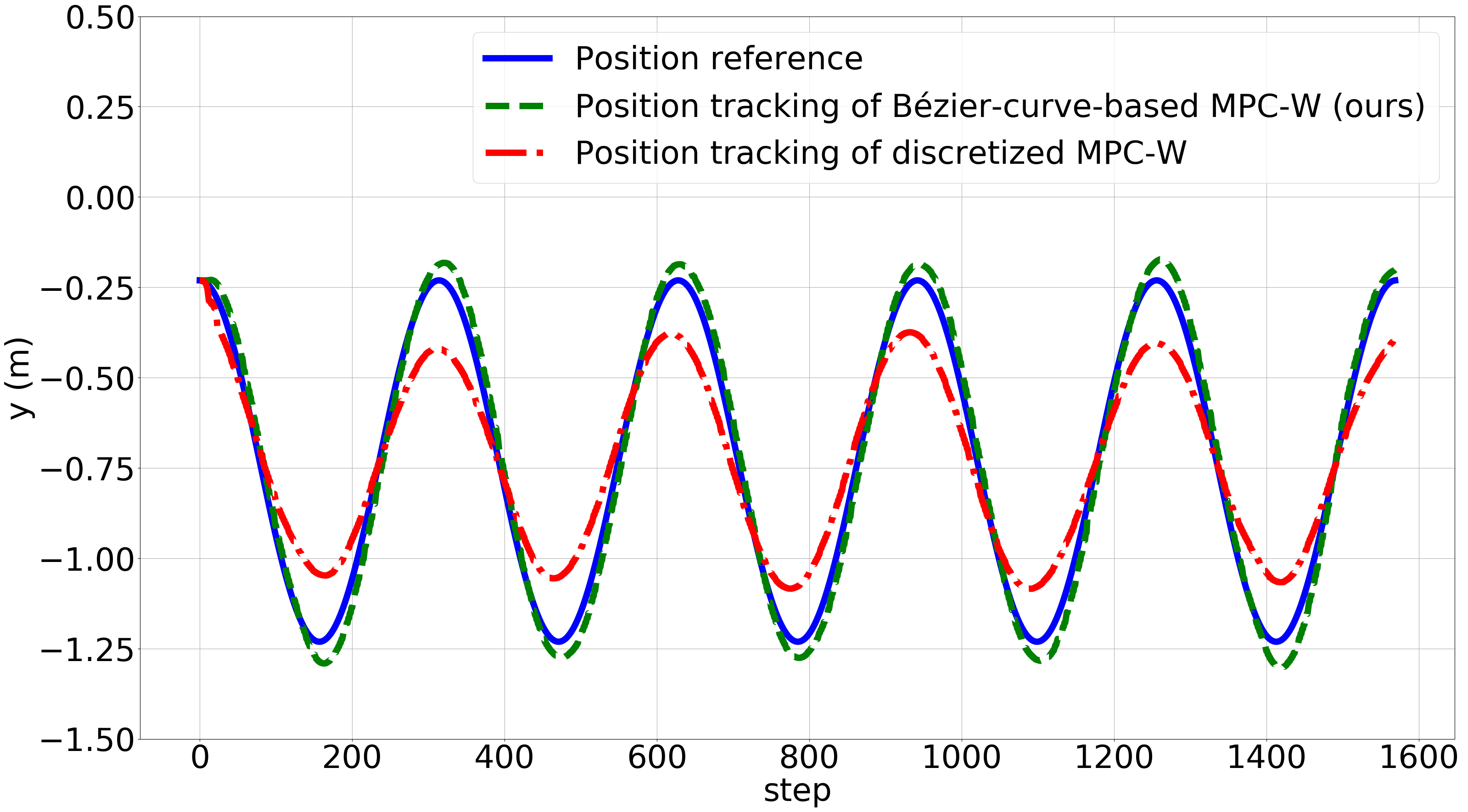}
	\caption{Tracking performance comparison of the right-arm end-effector position between discretized MPC-W and our B\'ezier-curve-based MPC-W using $K^{wb}=6$.
	}
	\label{simulation_comparison}
\end{figure}

\begin{figure}[ht]
	\centering
	\includegraphics[width = \linewidth]{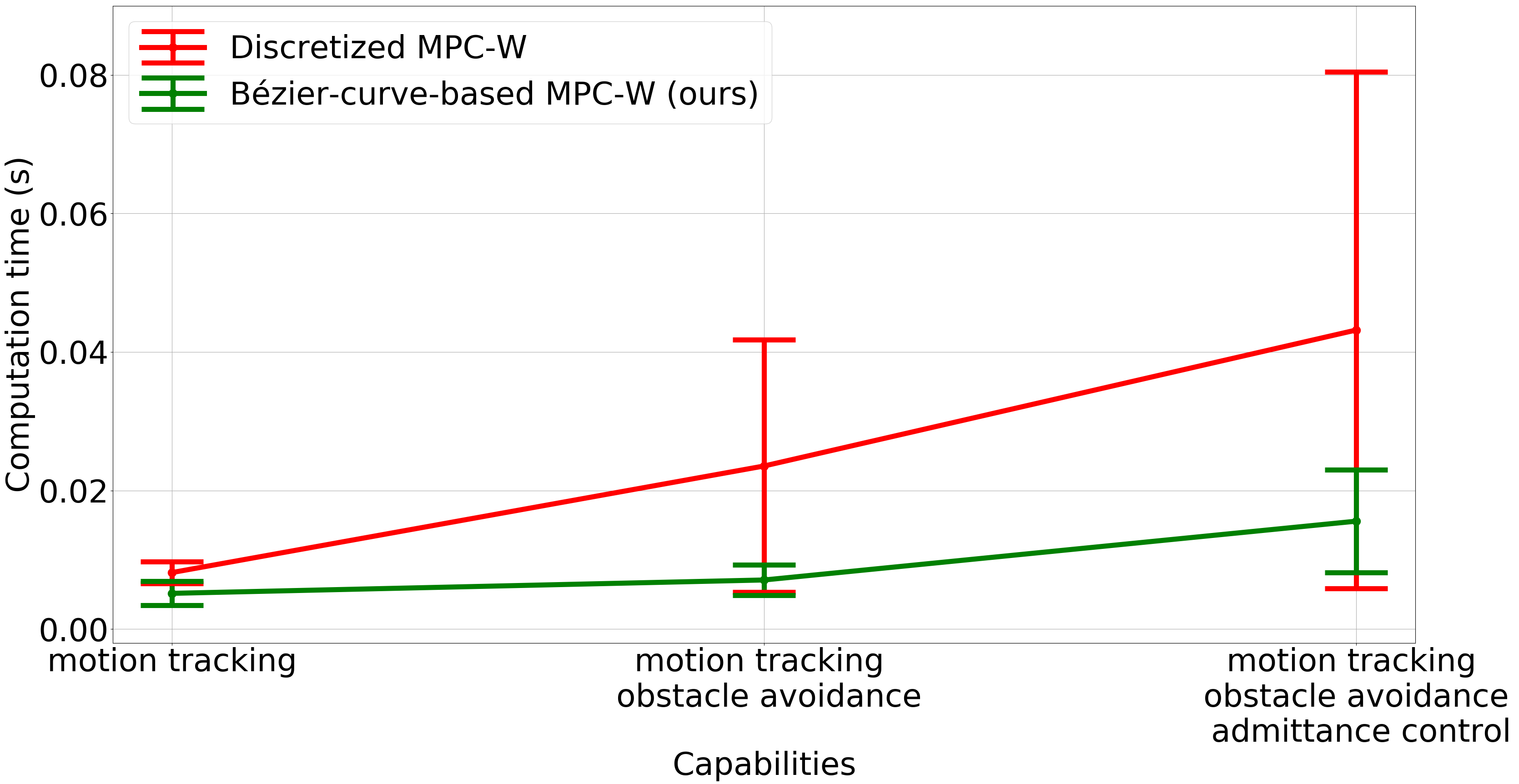}
	\caption{Comparison of computation times and standard deviations using different capabilities between discretization and our B\'ezier-curve-based whole-body MPC-W. For both methods, we set the same horizon ($1.25s$) and $6$ time knots. We set $6$ control points for all trajectories using our approach. 
	}
	\label{simulation_comparison_function}
\end{figure}

\begin{figure}[ht]
        \includegraphics[width = 0.93\linewidth]{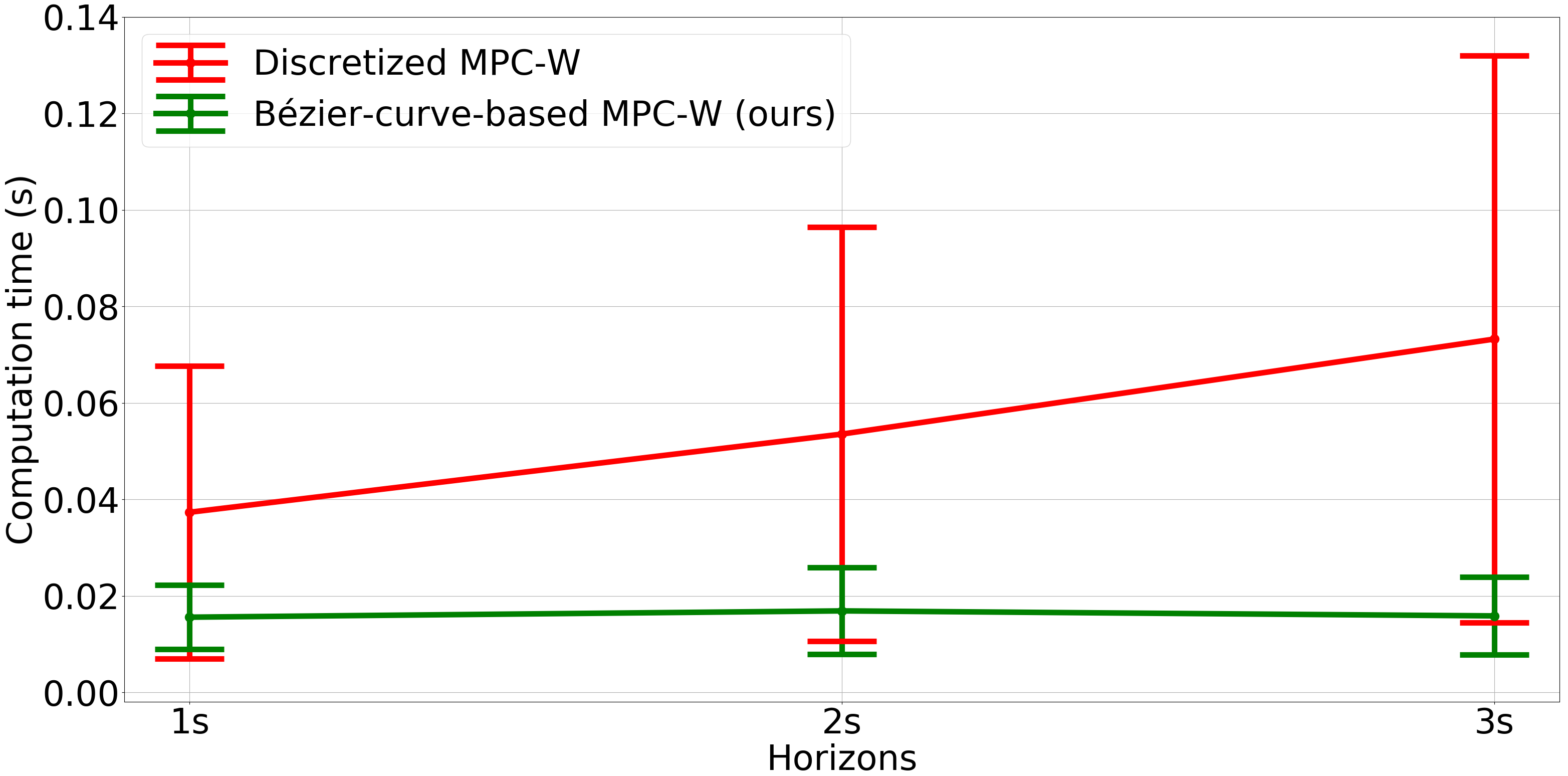}
	\caption{Comparative computation times and standard deviations using different horizons between discretization and our B\'ezier-curve-based whole-body MPC-W. $6$ time knots and control points are utilized. 
	}
	\label{simulation_comparison_horizon}
\end{figure}

\subsection{Fast Computation Speed of B\'ezier-curve-based MPC-T}\label{mpce_comparision_efficiency}
We compare three methods in terms of the computation efficiency of MPC-T in a simulated scenario in Fig. \ref{simulation_scenario}. These methods include the traditional discretization and B\'ezier-curve representation of the four quaternion components, and the B\'ezier-curve representation of $\bm{\psi}$ in \cref{subsec:stage1}. 

The results are shown in Tab. \ref{comparison_mpcE}. With unit-quaternion constraints, we can see that the direct discretization on $(\bm{p}, \bm{\theta})$
use the most compute time ($60.8ms$) for each control loop, with a standard deviation of $36.8ms$. The second method of representing $(\bm{p}, \bm{\theta})$ using B\'ezier curves, reduces the compute time to an average of $19.3ms$ and a standard deviation of $8.9ms$.
Even though this method significantly improves computation time, it does not ensure that all points on the trajectories are unit quaternions because the constraints are only enforced at specific time knots. 
Finally we compare our method, which represents the Bézier-curve-based quaternion trajectories on $\bm{\psi}$, ensuring the unit-quaternion property at all points. This removes the need for unit-quaternion constraints and results in a further reduction of MPC-T computation time to $6.3ms$ with a very small standard deviation of $2.4ms$.
This verifies the efficiency of our method in the first stage.

\begin{figure}[ht]
	\centering
	\includegraphics[height=5.5cm]{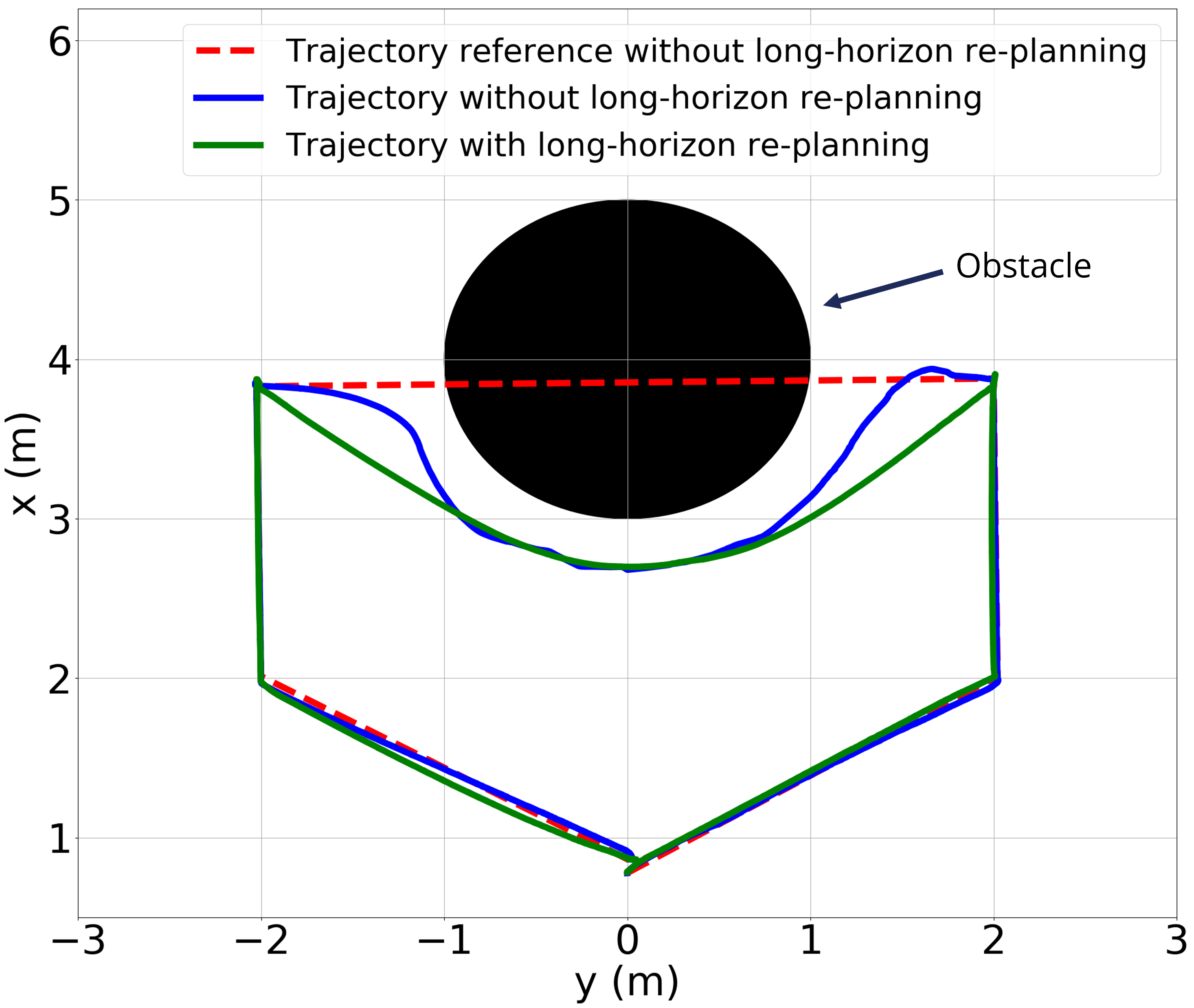}
	\caption{Comparison between a pre-defined trajectory and long-horizon re-planning using MPC-T. It shows the trajectories of the middle point of two end-effectors. 
	}
	\label{Curve_replanning}
\end{figure}

\begin{figure*}[t]
	\centering
 	\includegraphics[width=0.90\linewidth]{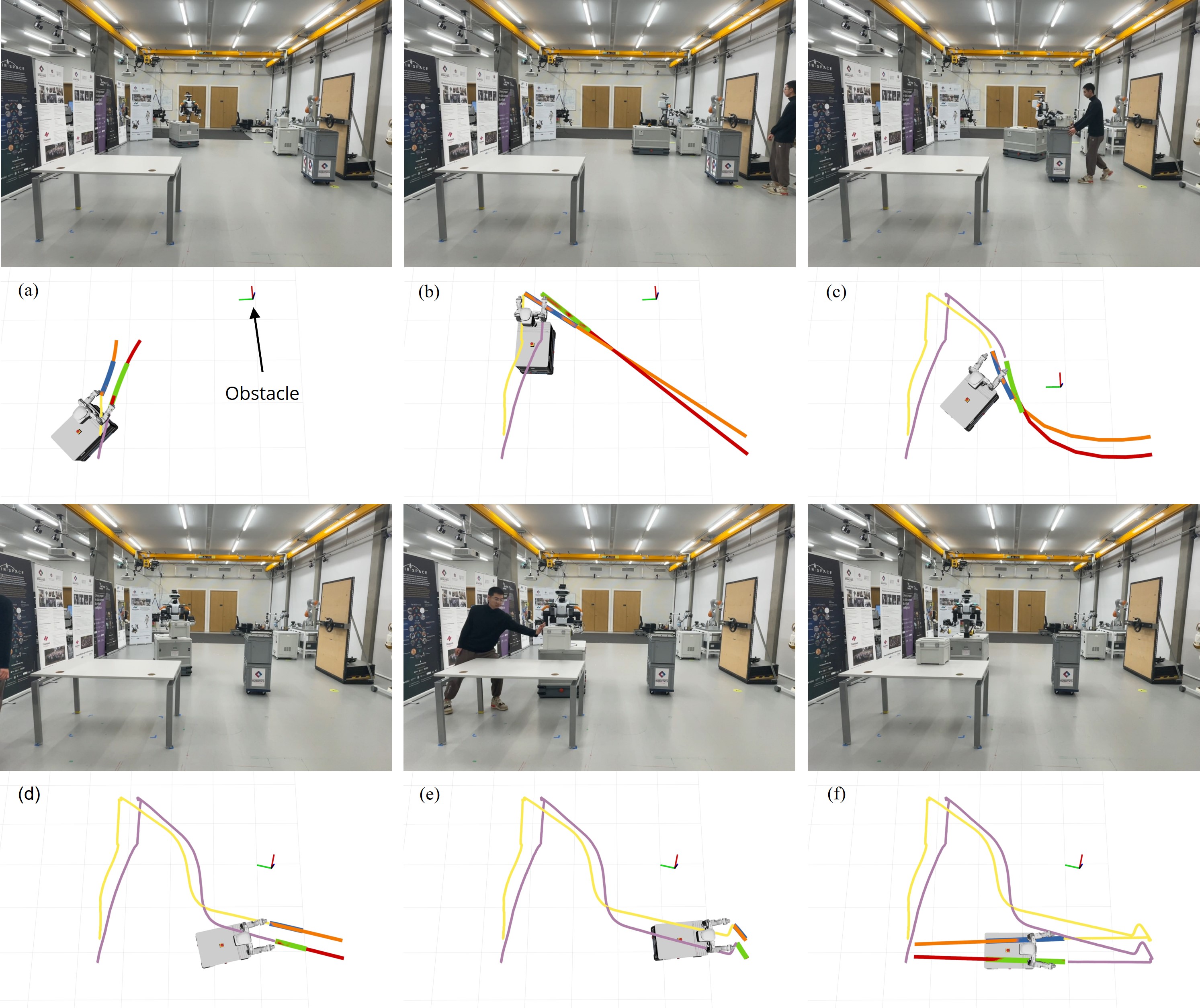}
	\caption{Experiment 1: Whole-body pick-and-place experiment with dynamic obstacle avoidance and compliant push-recovery motion. Purple and yellow lines denote the historical trajectories of the right and left end-effectors. Red and orange lines represent the motion plan of MPC-T from the current end-effector states to the goal. Green and blue lines stand for the motion plan of MPC-W in a relatively short horizon. (a) Approaching the box; (b, c, d) Executing dynamic obstacle avoidance; (e) Demonstrating push-recovery motion; (f) Returning to origin.}
	\label{EVA_experiment3}
\end{figure*}

\begin{figure}[ht]
	\centering
 	\includegraphics[width =0.95\linewidth]{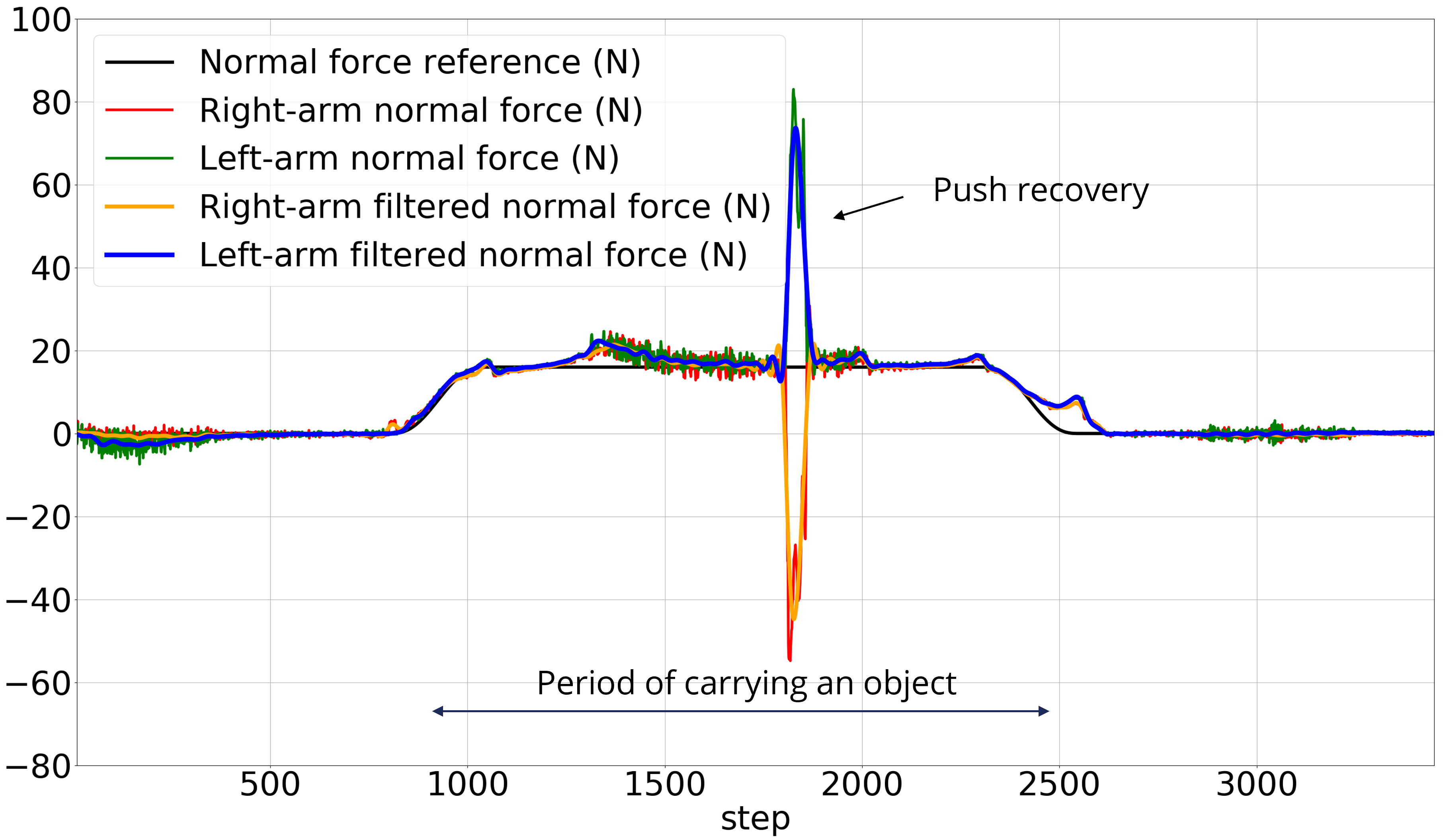}
	\caption{Normal force tracking for two arms in experiment 1.
	}
	\label{EVA_ft_experiment3}
\end{figure}

\begin{figure}[ht]
	\centering
 	\includegraphics[width=0.95\linewidth]{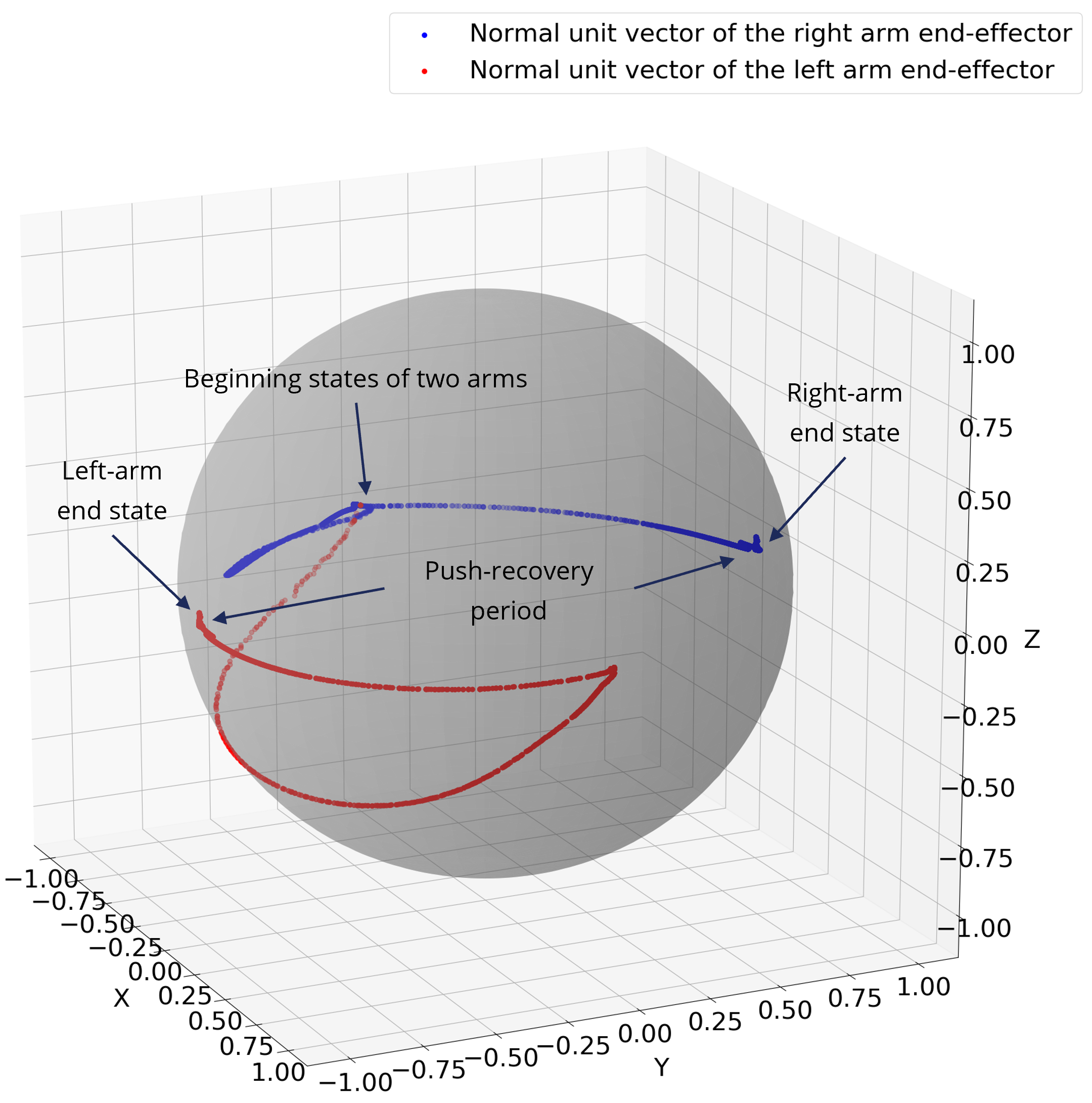}
	\caption{Quaternion trajectories of two end-effectors on a unit sphere in experiment 1, using the normal unit vectors of two end-effector palms. At the beginning, they are both $[-1, 0, 0]$.
	}
	\label{quaternion_experiment3}
\end{figure}

\subsection{Tracking Performance of B\'ezier-curve-based MPC-W}\label{tracking_performance}

We compare the computation and tracking performance of MPC-W between the discretization and B\'ezier-curve representation. Since the objectives of force and motion tracking conflict, admittance control is not added in this comparison to isolate its impact on the tracking performance.
We set the horizon to be $5s$ in both methods. Given a sine wave-shape position reference for two end-effectors, the tracking performance is shown in Fig. \ref{simulation_comparison}, and the comparison result is in Tab. \ref{comparison}. 
We can see that the tracking performance using the discretized MPC-W is relatively poor, with the mean absolute tracking error reaching $21.6\%$ when the number of time knots is small. Since the approximation of joint state transition is rough, the velocity command for the mobile base is not consistent with the position commands for the upper-body joints. To improve the tracking performance, dense time knots are required and we set the time knot number to be $26$. However, the computation time of discretized MPC-W becomes very slow, averaging $60.1ms$ with a large standard deviation of $35.7ms$. With the same long horizon and time knots, our Bezier-curve-based MPC-W runs faster and with better tracking performance, as well as a smaller standard deviation of computation time. 
Particularly, with dense time knots, our method still maintains a satisfactory tracking performance while running at a fast speed of $12.5ms$ with a standard deviation of $3ms$.

\subsection{Fast Computation Speed of B\'ezier-curve-based MPC-W for Different Capabilities and Horizons}\label{capabilities}

First, we look at the computational performance of MPC-W using the scenario in Fig. \ref{simulation_scenario} while addressing various tasks, including (a) motion tracking, (b) obstacle avoidance, and (c) force control -- using the discretized method and our B\'ezier-curve representation. 
The comparison result is shown in Fig. \ref{simulation_comparison_function}. Beginning from function (a), our method runs faster with an average computation time of $5.1ms$ compared to $8.1ms$ using the discretized MPC-W, while the standard deviations are similar to each other. Then we add the function (b) alongside (a), and we can see that our method is more competitive using only $7.1ms$, however, the computation time of discretized MPC-W increases almost 3 fold. The standard deviation obtained through our method is significantly smaller compared to that of the discretized MPC. Then the function (c) is induced, and our method only spends nearly one-third of the computation time of the discretized MPC-W. The standard deviation in our method is merely one-fifth of that observed in the discretized MPC-W. 
This comparison highlights the scalability of our method to diverse goals. 

\begin{figure*}[t]
	\centering
	\includegraphics[width=0.90\linewidth]{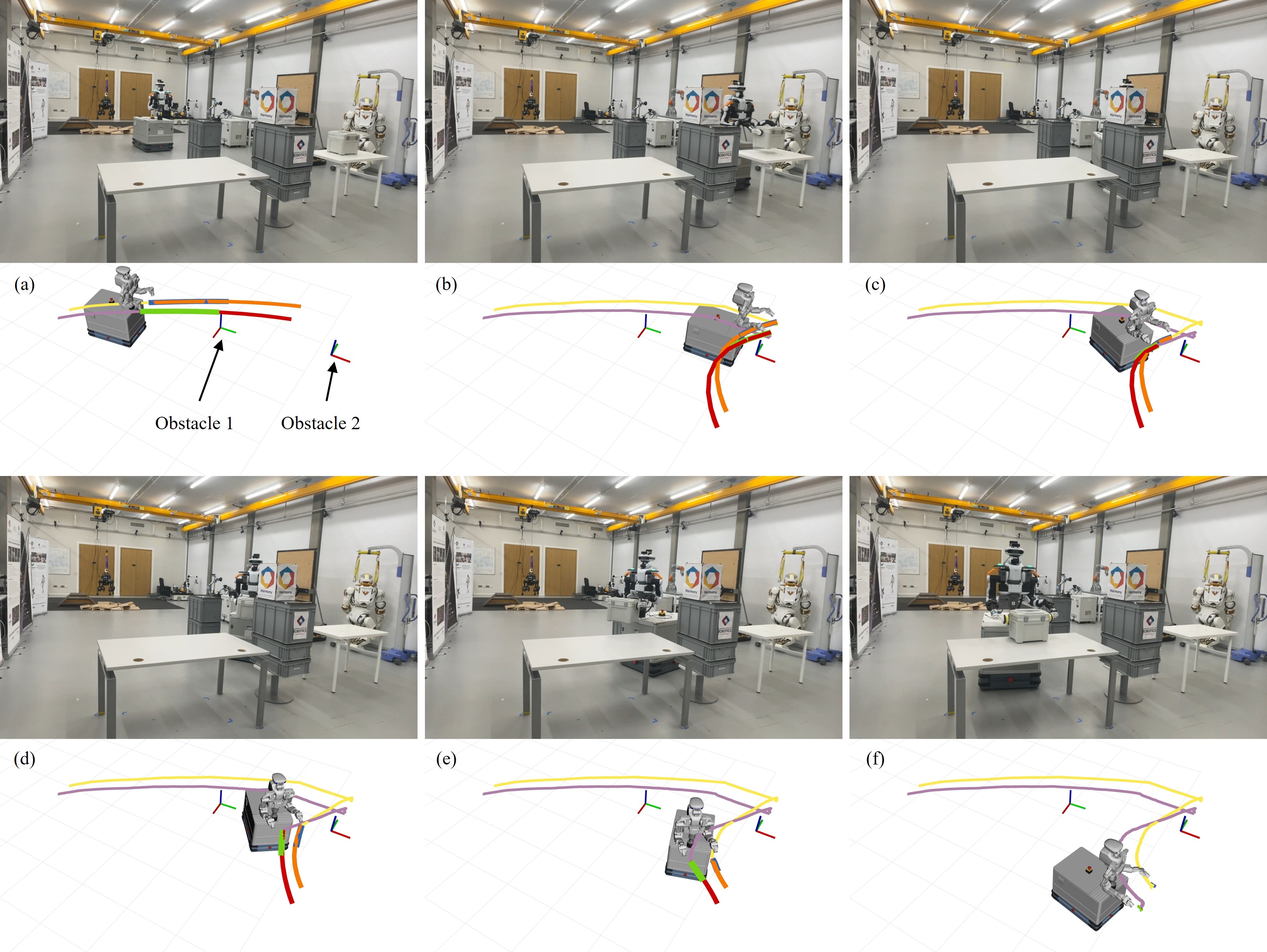}
	\caption{Experiment 2: Whole-body motion in narrow spaces. (a) Approaching the object while avoiding an obstacle; (b) Picking the object; (c, d, e) Transporting the object while navigating the narrow pathway; (f) Placing the object. 
	}
	\label{EVA_experiment2}
\end{figure*}

\begin{figure*}[ht]
	\centering
 	\hspace{-0.2cm}\includegraphics[width =0.84\linewidth]{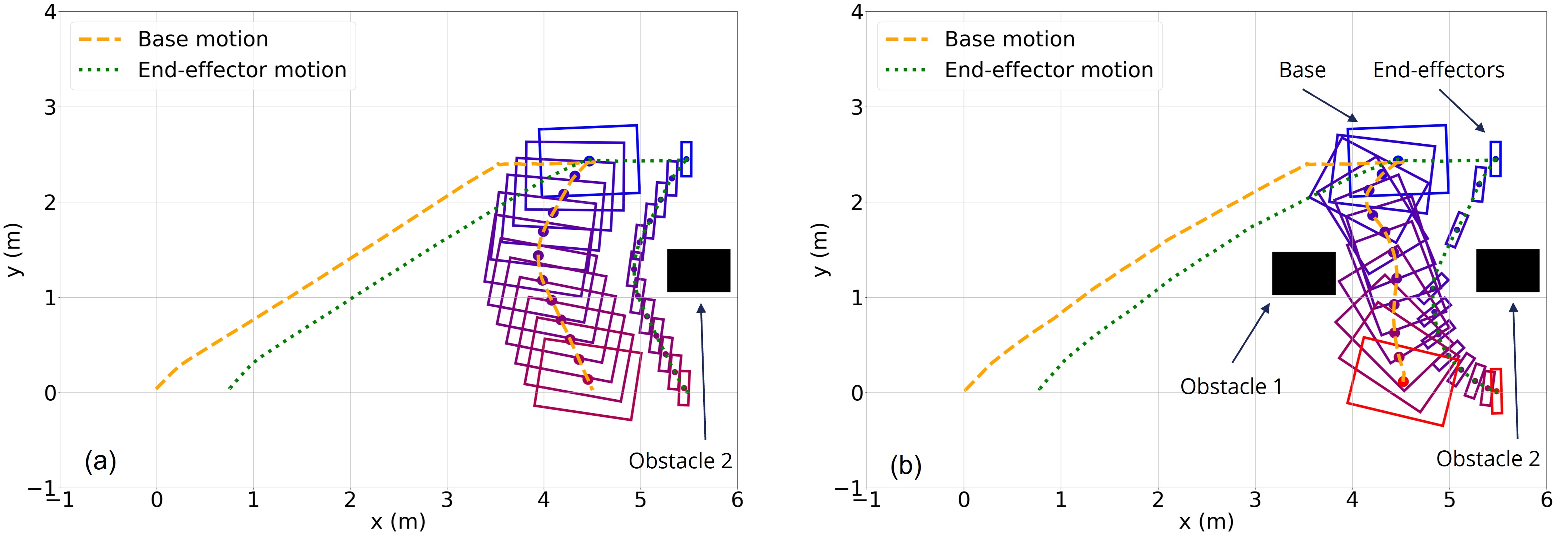}
	\caption{Base/end-effector motion in experiment 2. Big rectangles denote the base motion. Small rectangles cover the two effectors' motion. Black rectangles represent the two obstacles forming a narrow pathway that is navigated by the robot.
	}
	\label{EVA_base_arm_experiment2}
\end{figure*}

\begin{figure*}[t]
	\centering
	\includegraphics[width=0.90\linewidth]{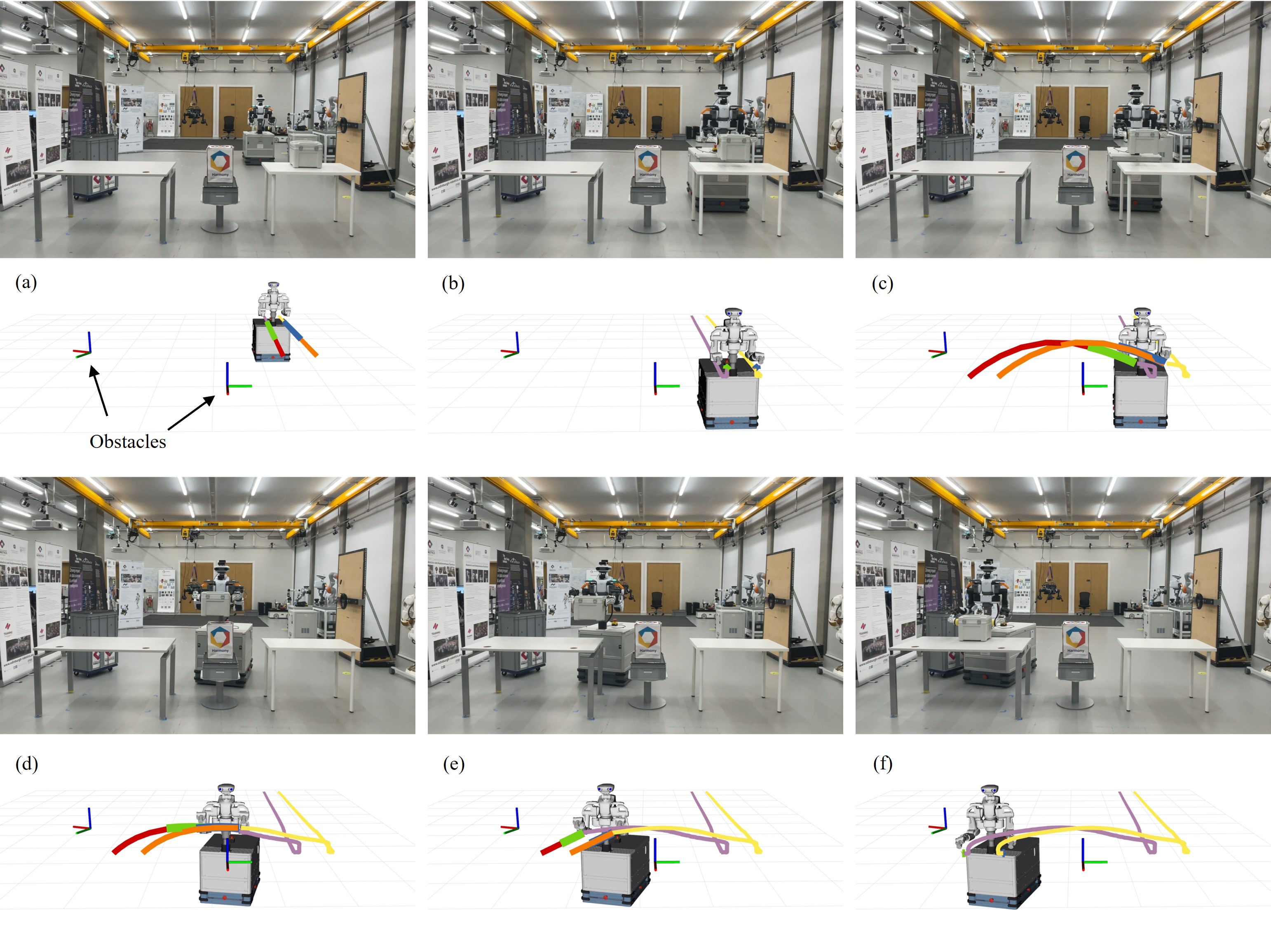}
	\caption{Experiment 3: Coordinated motion of lower and upper bodies while avoiding static obstacles. (a) Initial state; (b) Picking an object; (c)(d)(e) Transporting the object while avoiding an obstacle from its upper side; (f) Placing the object while twisting its base to avoid another obstacle. 
	}
	\label{EVA_experiment1}
\end{figure*}

Then we compare the two methods when handling different horizons ($1s$, $2s$, and $3s$) for all the functions. 
The results are shown in Fig. \ref{simulation_comparison_horizon}. Due to the coarse and inaccurate model-state transition in discretized MPC-W, it is evident that its computation times and standard deviations vary significantly across different horizons. In contrast, our method is more stable with significantly faster speeds and smaller standard deviations. 
Particularly, when the horizon is set to be at least $2s$, our MPC-W allows the robot to complete the entire mission in less than one-third of the time required by the discretized method.

\begin{figure}[t]
	\centering
 	\includegraphics[width =0.95\linewidth]{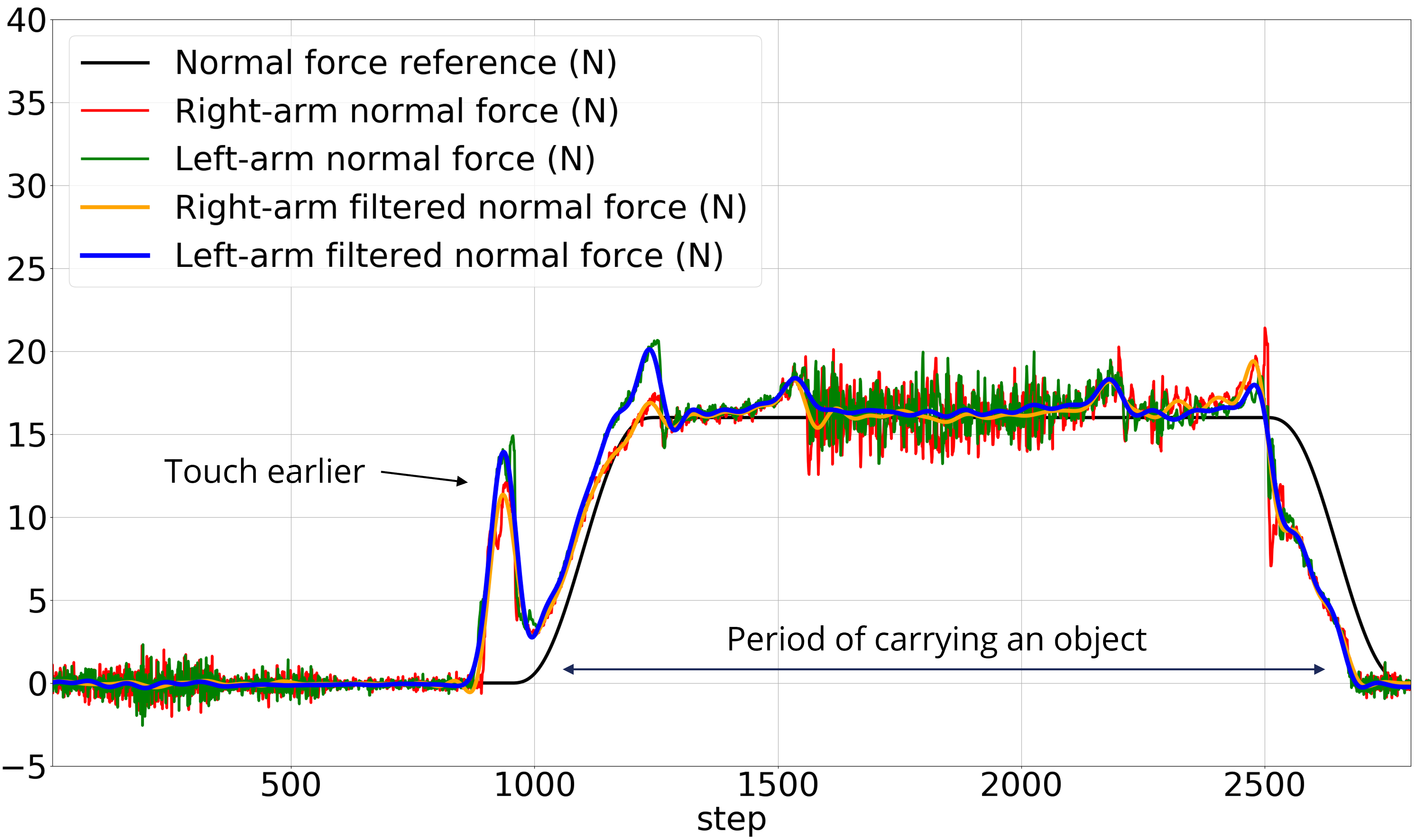}
	\caption{Normal force tracking for two arms in experiment 3.
	}
	\label{EVA_ft_experiment1}
\end{figure}

\subsection{Importance of an Effective Motion Reference for MPC-W}\label{mpce_comparision}

We compare the collision-avoidance performance of MPC-W between two different end-effector motion references using the scenario in Fig. \ref{simulation_scenario}. The MPC-W horizon is set to be $1.5s$ with $6$ time knots and control points. 

The first method applies a pre-defined and inappropriate end-effector trajectory (no update), and the obstacle avoidance only relies on MPC-W. To avoid the obstacle, the optimal solution of MPC-W is far from the motion reference and the average solving time is $31.4ms$ during obstacle avoidance. Without the first stage, the robot turns in front of the obstacle which is not natural or smooth, as shown in Fig. \ref{Curve_replanning}. 

The second method utilizes our re-planning module MPC-T to generate a collision-free trajectory in a long horizon, and the whole-body MPC-W tracks the online re-planned trajectory and avoids the obstacle in a more natural way, and the trajectory is much smoother, also shown in Fig. \ref{Curve_replanning}. Since MPC-T already generates a collision-free trajectory, it decreases the burden of MPC-W as the optimal solution in MPC-W is close to the re-planned motion reference, and MPC-W takes an average computation time of $21.8ms$ during obstacle avoidance. 
This verifies that compared to an inappropriate pre-defined motion plan, the online re-planning module MPC-T can generate a more suitable motion reference, accelerating the whole-body MPC-W computation speed and enabling it to produce smoother and more natural movements.

\subsection{Dynamic Obstacle Avoidance and Push Recovery}\label{experiment3}
In this experiment, we show the whole-body capabilities of transporting an object with a wide range of rotational motion and dynamic interaction with external disturbances, including dynamic obstacle avoidance and push recovery. 
The horizon for MPC-W is set to $2s$ and the longest horizon of MPC-T is $14s$ which occurs during the object-transport process. MPC-T and MPC-W utilize $8$ and $6$ time knots and control points, respectively, and this setting is also employed in the following experiments. The snapshots of this experiment are shown in Fig. \ref{EVA_experiment3}. 
MPC-T and MPC-W spend an average of $5.4ms$ and $13.3ms$ for each control loop, with the standard deviations of $1.9ms$ and $5.4ms$, respectively. Fig. \ref{EVA_experiment3} shows the historical trajectories of two end-effectors, the planned long-horizon trajectories from MPC-T, and the generated tracking trajectories from MPC-W. We can see that the tracking performance of MPC-W is quite satisfactory.

When the robot begins to transport the object, the end-effector motion plan is shown in Fig. \ref{EVA_experiment3}(b). Then we push an obstacle towards the robot. Due to the fast speeds of two MPCs, the robot is able to re-plan the end-effectors' motion efficiently using MPC-T and generate whole-body motion online via MPC-W, facilitating avoiding the dynamic obstacle with a smooth path, as shown in Fig. \ref{EVA_experiment3}(c, d). When the robot almost arrives at the target table, we push the right-arm end-effector. The normal forces of two end-effectors expressed in their local frames are shown in Fig. \ref{EVA_ft_experiment3}. The robot performs compliant motion efficiently, and after the push, the robot recovers and handles the remaining tasks, shown in Fig. \ref{EVA_experiment3}(e). Note that in MPC-T, we only set the final goal of each phase, and the relative distance between two end-effectors cannot be ensured to be equal to the object width. However, to track the force reference, the integrated admittance control in MPC-W enables the robot to perform stable and compliant contact with the object. During this push recovery, the translational motion of end-effectors is shown in Fig. \ref{EVA_experiment3}(e), and their rotational motion is depicted on a unit sphere in Fig. \ref{quaternion_experiment3}. Under the dynamic environment and external disturbances, the reactive trajectories are adaptive and smooth in an online fashion.   

\subsection{Navigation in Narrow Spaces}\label{experiment2}
In this experiment, we show that MPC-W is able to correct the generated end-effectors' motion from MPC-T.
As shown in Fig. \ref{EVA_experiment2}, the robot's mission is to transport an object while passing a narrow space. Since a relatively long horizon is important for the robot to pass the narrow path in a smooth way, the horizon of MPC-W is set to be 3s and the longest horizon of MPC-T is $15s$. 
The motion of the mobile base and two end-effectors is depicted in Fig. \ref{EVA_base_arm_experiment2}. 

We first test the scenario with only obstacle 2, the base does not require the function of obstacle avoidance or undergo significant turning, as shown in Fig. \ref{EVA_base_arm_experiment2}(a). However, when two obstacles are configured, as shown in Fig. \ref{EVA_experiment2} and Fig. \ref{EVA_base_arm_experiment2}(b), the limited width of the path is not enough for EVA to maneuver sideways.
Since MPC-W incorporates obstacle avoidance between the mobile base and the two obstacles, it corrects the MPC-T trajectory considering whole-body motion constraints and forces the entire robot to pivot before navigating the narrow path, as shown in Fig. \ref{EVA_base_arm_experiment2}(b). The interplay between MPC-T and MPC-W results in a dynamic change in the robot's motion.
The average computation times for MPC-T and MPC-W are $5.5ms$ and $13.2ms$, and the standard deviations are $1.7ms$ and $7.8ms$, respectively.
This experiment validates the capability of MPC-W to rectify the generated trajectories from MPC-T efficiently, to ensure the robot safety in the narrow space. 

\begin{figure*}[t]
	\centering
	\includegraphics[width=0.90\linewidth]{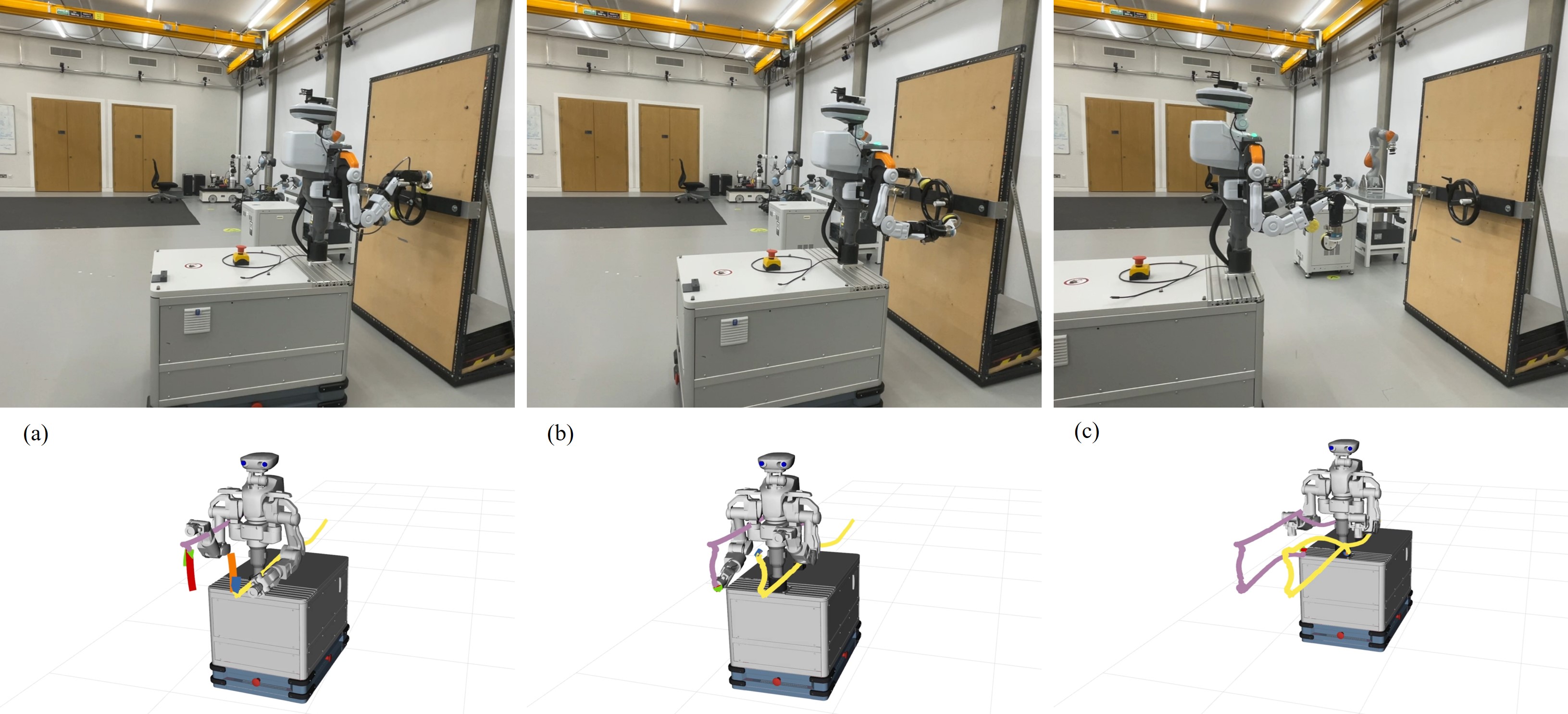} 
	\caption{Experiment 4: Dual-arm manipulation with object shape adaption. (a) shows the unreasonable motion plan for wheel turning from MPC-T. (b) depicts the adaptive motion for wheel turning. (c) shows the whole-process trajectory.
	}
	\label{EVA_experiment4}
\end{figure*}

\subsection{Coordinated  Lower and Upper Body Movements}\label{experiment1}
In this experiment, we test the coordinated motion generation of lower and upper bodies.
The cost function relating to minimizing the dual-arm motion with a suitable weight enables the dual-arm end-effectors to be close to the mobile base.
As shown in Fig. \ref{EVA_experiment1}, EVA transports a box from the right table to the left one. In the middle of two tables, an obstacle is placed which forces EVA to lift up two arms to a big range automatically. Another obstacle is placed near the left table which forces EVA to twist its mobile base while placing the box at the target. In this scenario, the longest horizon $13s$ of MPC-T happens when the robot transports the object between two tables. The horizon for MPC-W is $2s$. The average computation time of MPC-T and MPC-W is $5.5ms$ and $14.6ms$, respectively, and their standard deviations are $1.8ms$ and $5.8ms$, correspondingly.
In addition, the normal forces of two end-effectors are shown in Fig. \ref{EVA_ft_experiment1} and the force-tracking performance is satisfactory. Even though the end-effectors touch the object earlier, the integrated admittance controller promptly adjusts the end-effector motion, facilitating efficient tracking of the force reference.

\subsection{Dual-arm Manipulation with Object Shape Adaptation }\label{experiment4}
As shown in \Cref{EVA_experiment4}, we test our framework for dual-arm manipulation with object-shape adaptation since the manipulated wheel shape is unknown to the robot. MPC-W horizon is set to be $2s$ and the longest horizon of MPC-T is $8s$ which corresponds to the turning-wheel period.
When two end-effectors reach the starting-turning pose, the planned trajectories via MPC-T are depicted in Fig. \ref{EVA_experiment4}(b) which resemble almost two straight lines. Even though the trajectories are not reasonable for the round-shaped wheels, our bilevel-MPC framework is robust and compliant to handle this uncertainty. 
Then adaptive turning motion is generated, and the actual end-effectors' trajectories are shown in Fig. \ref{EVA_experiment4}(c) which are consistent with the wheel shape. 
With and without admittance control, the maximum normal force is $40N$ and $144N$, respectively.
The interplay of MPC-T and MPC-W incorporating admittance control 
enhances the manipulation adaption, solid contact, and safety between the end-effectors and the manipulated object. 
The average computation time of MPC-T and MPC-W for each control loop is $5.2ms$ and $11.6ms$, with standard deviations of $2.2ms$ and $2.5ms$, respectively.  

\section{Conclusion} \label{conclusion}
This paper introduces an efficient representation to address online whole-body motion planning of bi-manual mobile manipulation within our bilevel-MPC framework. 
We integrate B\'ezier-curve parameterization into two MPCs to avoid inaccurate model-state transition and reduce the decision variable number. This efficient representation enables the task-space MPC-T to realize fast motion planning of two collaborative end-effectors' trajectories in $SE(3)$ in a long horizon while considering the mobile base motion limits. Additionally, it facilitates the whole-body MPC-W to generate whole-body joint motion online while satisfying various hard constraints. With this efficient representation, we also achieve online predictive admittance control for our robot with high DOFs. In each control loop, MPC-T updates a motion reference that speeds up the convergence of MPC-W, while MPC-W follows and corrects the generated trajectory from MPC-T. Our approach shows strengths in online whole-body motion planning and the generation of consistent position/velocity commands for our highly redundant dual-arm mobile manipulator. Consequently, 
the bilevel-MPC framework empowers our robot to navigate various obstacles efficiently, respond compliantly to external disturbances, exhibit robustness, and generate adaptive trajectories facing uncertainties.

\section*{Acknowledgment}
This research is supported by the EU H2020 projects Enhancing Healthcare with Assistive Robotic Mobile Manipulation (HARMONY, 101017008), the Kawada Robotics Corporation, and The Alan Turing Institute.

\bibliographystyle{IEEEtranBST/IEEEtran}
\bibliography{manuscript} 

\end{document}